\documentclass[a4paper,oneside,12pt]{article}

\usepackage{microtype}
\usepackage{enumitem}
\usepackage{pdflscape}
\usepackage{booktabs}
\usepackage{graphicx}
\usepackage[subrefformat=parens,labelformat=parens, position=top]{subfig}
\usepackage[authoryear]{natbib}
\setcitestyle{authoryear}

\usepackage{url}

\usepackage[nottoc]{tocbibind}

\usepackage{amsmath,amssymb,amsthm}
\usepackage{mathtools}
\usepackage{gensymb}
\usepackage{bm}

\usepackage{pdflscape}
\usepackage[title]{appendix}

\newtheorem{theorem}{Theorem}
\newtheorem{definition}{Definition}

\newcommand{\R}{{\rm I\!R}}

\newcommand{\Prob}{{\rm I\!P}}
\newcommand{\Exp}{{\mathop{\rm I\!E}}}

\newcommand{\norm}[1]{\left\lVert#1\right\rVert}

\DeclarePairedDelimiterX{\infdivx}[2]{(}{)}{%
	#1\;||\;#2%
}

\DeclarePairedDelimiterX{\inner}[2]{\langle}{\rangle}{#1, #2}

\usepackage[colorlinks=true,allcolors=blue]{hyperref}

\title{Normality Testing with Neural Networks}
\author{Milo{\v s} Simi{\' c}\\University of Belgrade\\Studentski trg 1, 11000 Belgrade\\milos.simic.csci@gmail.com\\milos.simic.ms@afrodita.rcub.bg.ac.rs}
\date{}

\begin{document}
	\maketitle
	\abstract{In this paper, we treat the problem of testing for normality as a binary classification problem and construct a feedforward neural network that can successfully detect normal distributions by inspecting their small samples. The numerical experiments conducted on the samples with no more than $100$ elements indicated that the neural network which we trained was  more accurate and far more powerful than the most frequently used and most powerful standard tests of normality: Shapiro-Wilk, Anderson-Darling, Lilliefors and Jarque-Berra, as well as the kernel tests of goodness-of-fit. The neural network had the AUROC score of almost $1$, which corresponds to the perfect binary classifier. Additionally, the network's accuracy was higher than $96\%$ on a set of larger samples with $250-1000$ elements. Since the normality of data is an assumption of numerous statistical techniques, the neural network constructed in this study has a very high potential for use in everyday practice of statistics, data analysis and machine learning in both science and industry. AAA}
	\section{Introduction}\label{sec:introduction}
	An underlying assumption of many statistical procedures is that the data at hand come from a normal distribution \citep{Razali2011,Thode2002testing}. Hence, normality tests are among the most frequently used statistical tools.
	
	The standard statistical tests of normality require two hypotheses to be formulated: the null ($H_0$) that the data at hand come from a normal distribution, and the alternative hypothesis ($H_a$) that the distribution from which the data were drawn is not normal. The tests are usually conducted as follows. First, researchers choose the value of $\alpha$, the probability to incorrectly reject a true $H_0$. The most common values for $\alpha$ are $0.01$ and $0.05$. Then, the test statistic is calculated and its $p$-value is determined. If the $p$-value is lower than $\alpha$, the null hypothesis is rejected, and the data are not considered normal. Otherwise, $H_0$ is not rejected. The choice of $\alpha$ limits the Type I error rate, i.e. the frequency at which the test rejects a true $H_0$.
	
	Each statistical test for normality is based on a statistic that summarizes the sample being analyzed and considers only a specific property of normal distributions \citep{Razali2011}, which is why different tests sometimes give conflicting results \citep{Sigut2006}. Moreover, numerous simulation studies have shown that the tests exhibit low power (probability to identify a non-normal distribution) when applied to samples with fewer than $100$ elements \citep{Esteban2001,Noughabi2011,ComparisonThesis2010,Razali2011,YapSim2011,Ahmad2015,Marmolejo2013,Mbah2015,Yusoff2012,Patricio2017,Wijekularathna2019}. This poses a problem when only a small sample is available. In such cases, failure to detect non-normality and carrying out a procedure, such as the F test (used in ANOVA), that assumes that the data follow a normal distribution but is sensitive to departures from normality, can give extremely misleading answers and lead to wrong conclusions \citep{Wilson1990}. In this study, we propose a new test of normality, based on artificial neural networks, with the aim of making it more accurate on small samples than the standard tests of normality. Our research hypothesis is that neural networks can learn efficient classification rules from the datasets of normal and non-normal samples and that they can be used to classify a distribution in question as normal or non-normal based on a small-size sample drawn from it. In our experiments, it turned out that the neural networks outperformed the standard tests of normality. To our best knowledge, this is the third attempt at testing for normality with neural networks, the previous two being the studies of \cite{Wilson1990} and  \cite{Sigut2006}, whose work we build on. Another two contributions are: (1) introduction of descriptors, the data structures which allow machine-learning algorithms to be applied to variable-size and set data, and (2) creation of large and diverse datasets of samples drawn from various distributions.

	The rest of the paper is organized as follows. We review the literature in Section \ref{sec:literature_review}. In Section \ref{sec:tests_of_normality}, we formally define the problem of normality testing and briefly sketch a number of selected tests of normality. In Section \ref{sec:neural_networks}, we describe how we constructed neural networks that can classify samples as coming from normal or non-normal distributions. The datasets used are presented in Section \ref{sec:datasets}. The results of evaluation are given in Section \ref{sec:evaluation} and  discussed in Section \ref{sec:discussion}. The conclusions are drawn in Section \ref{sec:conclusion}.
	
	Throughout the paper, we use the common mathematical notation: $\Prob$ denotes an unnamed probability measure, $\Exp$ the expectation of a random variable, the uppercase Latin letters represent random variables, $\bm{1}\{\varphi\}$ is the indicator function equal to $1$ when the condition $\varphi$ is true, and $0$ otherwise, and so on. The cumulative distribution function (CDF) of the standard normal distribution $N(0, 1)$ is denoted as $\Phi$. EDF is the abbreviation for the empirical distribution function. For a sample $\mathbf{x}=[x_1,x_2,\ldots,x_n]$, $\overline{x}$ represents the sample's mean and $\text{sd}(\mathbf{x})$ the usual estimate of its standard deviation.
	\section{Literature Review}\label{sec:literature_review}
	Machine learning algorithms have been applied to several problems which were traditionally in the domain of statistics. In this Section, we review the approaches taken by machine-learning researchers in solving those problems.
	
	In  the \textbf{the two-sample problem}, the goal is to decide if two samples of data have been drawn from the same distribution. If the elements of the two samples are treated as the objects of two classes, then a binary classifier can be trained to discriminate between the two samples. The classifier's accuracy can then serve as a statistic upon which a new statistical test can be based. In the case that the samples have been drawn from the same distribution, we expect low accuracy, and vice versa. This approach was followed in several studies, e.g.: \cite{LopezPaz2017,Ojala2010,AlRawi2012,Kim2016,Blanchard2010,Rosenblatt2019}. In some of them, the new test statistic is defined to be the training accuracy score, whereas others split the samples into training and test subsets and use the test accuracy as the statistic. Kernel tests constitute another group of machine-learning tests for the two-sample problem. \cite{Gretton2012} continue the research of \cite{Borgwardt2006,Gretton2007a,Gretton2007b,Smola2007,Gretton2009} and define non-parametric tests using kernels and reproducing kernel Hilbert spaces (RKHS) \citep{Scholkopf2001,Steinwart2008}. The kernel tests are based on the fact that for a convenient choice of kernel, each distribution can be mapped into a unique point in the kernel's RKHS, which is called the distribution's embedding. The expected distance between the samples' embeddings is zero if and only if they are from the same distribution. That distance is calculated using kernels.
	
	In \textbf{the statistical independence problem}, one is interested in finding out if two or more random variables are statistically independent. \cite{Gretton2008} formulate a statistical test of independence of two variables based on the Hilbert-Schmidt Independence Criterion (HSIC) proposed by \cite{Gretton2005}, which is a measure of statistical independence. HSIC is also used by \cite{Chwialkowski2014} to check if two time series are independent. Similarly, \cite{Pfister2017} propose HSIC-based tests of independence of an arbitrary number of random variables.
	
	Finally, \textbf{the goodness-of-fit problem} is the one in which we check if a sample of data has been drawn from one particular distribution of interest, which we call a model. \cite{Chwialkowski2016} define the model's Stein operator $T_q$ \citep{Stein1972} over an RKHS. For any function $f$ in the RKHS, the expected value of $(T_qf)(X)$ is zero if and only if $X$ follows the model distribution. \cite{Chwialkowski2016} show that $(T_qf)(X)$ is a measure of discrepancy between $X$ and the model, call it the Stein discrepancy and calculate it using kernels. However, the asymptotic distribution of its normalized estimator in the case of a fit has no computable closed form \citep{Chwialkowski2016}, so \cite{Chwialkowski2016} use wild bootstrap \citep{Shao2010,Leucht2013} to address the issue and account for possible correlations in the data. The same idea is independently studied by \cite{Liu2016}. \cite{Liu2016} define a different estimator of the Stein discrepancy. The estimator's distribution, in the case that the data are not generated by the model, is asymptotically normal, but in the case that they indeed were drawn from the model, the limiting distribution of the estimator has no analytical form. \cite{Liu2016} estimate it by the bootstrap method proposed by \cite{Arcones1992,Huskova1993}. \cite{Wittawat2017} follow the approach of \cite{Chwialkowski2015} for kernelized two-sample tests and present a variant of the Stein discrepancy statistic that is comparable to the original in terms of power, but can be computed faster. We describe the test of \cite{Wittawat2017} in more detail in Section \ref{sec:tests_of_normality}, along with other normality tests with which we compared our network.
	
	\cite{Lloyd2015,Kellner2019} show how the kernel-based tests can be used to check if a sample comes from a normal distribution. \cite{Lloyd2015} fit a Gaussian model to the histogram of Simon Newcomb’s 66 measurements taken to determine the speed of light. Then, they simulate a thousand points from the fit model and an carry out a kernel two-sample test on the original and simulated data. \cite{Kellner2019} estimate the normal model's mean and variance from the sample, which gives a specific normal distribution and its corresponding embedding in the chosen kernel's RKHS. The squared distance between that point and the sample's embedding is easily computed via the kernel trick. \cite{Kellner2019} find the ($1-\alpha$)-quantile of the squared distance estimator's distribution via the fast bootstrap method of \cite{Kojadinovic2012} and thus determine the test's rejection region. This is equivalent to rejecting the hypothesis that the data are normal if the $p$-value of the test statistic is lower than $\alpha$ ($\alpha\in(0,1)$).
	
	\cite{Wilson1990} are the first to test for normality with neural networks. They train a neural network on a dataset of samples consisting from the standard normal and selected non-normal distributions, each sample having exactly $30$ elements. Before training the network, \cite{Wilson1990} map each sample to a vector of $16$ statistics (which are not specified in the paper). \cite{Sigut2006} adopt a similar approach. They train a neural network on a dataset consisting of many normal and non-normal samples drawn from a large number of different distributions using the Johnson system of distributions \citep{Johnson1949a,Johnson1949b}. In the study of \cite{Sigut2006}, samples can contain any number of elements, which requires a preprocessing step to transform the samples to the vectors of the same space. \cite{Sigut2006} map each sample to a vector of several statistics: skewness, kurtosis, the $W$-statistic of the Shapiro-Wilk test of normality  \citep{Shapiro1965}, the statistic of the Lin-Mudholkar test \citep{Lin1980}, the statistic of the Vasicek entropy test of normality \citep{Vasicek1976}, and the sample size. The idea central to the approaches of \cite{Wilson1990} and \cite{Sigut2006} is  that the neural networks can use the insights all those different statistics give into a sample and outperform the tests based on individual statistics only. The network of \cite{Sigut2006}  is presented  in more detail in Section \ref{sec:tests_of_normality}.
	
	\section{Tests of Normality}\label{sec:tests_of_normality}
	The problem of testing for normality can be formulated as follows. 
	\begin{definition}\label{def:testing_for_normality}
		Given a sample $\mathbf{x}=[x_1, x_2, \ldots, x_n]$ drawn from a distribution modeled by a random variable $X$, choose between the following two hypotheses:
		\begin{itemize}
			\item the null ($H_0$) that $X$ follows a normal distribution ($X \sim N$);
			\item the alternative hypothesis ($H_a$) that $X$ does not follow a normal distribution ($X \not \sim N$).
		\end{itemize}
	\end{definition}
	The problem can also be formulated for the case in which we want to test if $X$ follows a specific normal distribution $N(\mu, \sigma^2)$. However, we will focus on the formulation given above because it represents a more realistic scenario. In all that follows, we assume that the distributions are one-dimensional.
	
	There are two main types of techniques for normality testing:
	\begin{itemize}
		\item \textbf{graphical methods}, which provide qualitative assessment of normality of the sample in question \citep{Henderson2006}: histograms, box--and--whisker plots, quantile--quantile (Q--Q) plots, probability--probability (P--P) plots;
		\item \textbf{numerical methods}, which provide quantitative assessment of normality.
	\end{itemize}
	
	The methods in the latter group can be divided into four different categories based on the characteristic of the normal distribution that they focus on \citep{Seier2011}:
	\begin{enumerate}
		\item tests based on skewness and kurtosis;
		\item tests based on on the EDF;
		\item regression and correlation tests;
		\item other tests for normality.
	\end{enumerate}
	
	There are forty classical statistical tests for normality \citep{Dufour1998}. We compare our network with four of them: the Shapiro--Wilk (SW) \citep{Shapiro1965}, Liliefors (LF) \citep{Lilliefors1967}, Anderson-Darling (AD) \citep{Anderson1952,AndersonDarling1954}, and Jarque--Bera (JB) tests \citep{JarqueBera1980,Jarque1987}. We chose those tests because they are the most frequently examined in simulation studies and are the most powerful tests of normality. Of the new tests, we included the kernel tests of normality \citep{Wittawat2017}, since they also bridge machine learning and statistics as our approach, but in another way. Finally, we included the network of \cite{Sigut2006} in the study as it is comparable and complementary to ours. 
	
	Throughout this Section, $\mathbf{x}=[x_1, x_2, \ldots, x_n]$ will denote a sample drawn from the distribution whose normality we want to test.
	
	\subsection{The Shapiro-Wilk Test}\label{subsec:shapiro-wilk}
	
	Let $\mathbf{b} = [b_1, b_2, \ldots, b_n]^T$ denote the vector of the expected values of order statistics of independent and identically distributed random variables sampled from the standard normal distribution. Let $\mathbf{V}$ denote the corresponding covariance matrix.
	
	The intuition behind the SW test is as follows. If a random variable $X$ follows normal distribution $N(\mu,\sigma^2)$, and $Z\sim N(0, 1)$, then $X=\mu+\sigma Z$ \citep{Shapiro1965}. For the ordered random samples $\mathbf{x}=[x_1,x_2,\ldots,x_n]\sim N(\mu, \sigma^2)$ and $\mathbf{z}=[z_1,z_2,\ldots,z_n]\sim N(0,1)$, the best linear unbiased estimate of $\sigma$ is \citep{Shapiro1965}:
	\begin{equation}
	\hat{\sigma}=\frac{\mathbf{b}^T\mathbf{V}^{-1}\mathbf{z}}{\mathbf{b}^T\mathbf{V}^{-1}\mathbf{b}}
	\end{equation}
	In that case, $\hat{\sigma}^2$ should be equal to the usual estimate of variance $\text{sd}(\mathbf{x})^2$:
	\begin{equation}
	\text{sd}(\mathbf{x})^2=\frac{1}{n-1}\sum_{i=1}^{n}(x_i-\overline{x})
	\end{equation}
	The value of the test statistic, $W$, is a scaled ratio of those two estimates:
	\begin{equation}\label{eq:w-statistic}
	W = \frac{\left(\sum_{i=1}^{n}a_ix_i\right)^2}{\sum_{i=1}^{n}\left(x_i - \overline{x}\right)^2}
	\end{equation}
	where:
	\begin{equation}
	\textbf{a} = [a_1, a_2, \ldots, a_n]  =  \frac{\textbf{b}^T\textbf{V}^{-1}}{\left({\textbf{b}^T\textbf{V}^{-1}\textbf{V}^{-1}\textbf{b}}\right)^{1/2}}
	\end{equation}
	The range of $W$ is $[0,1]$, with higher values indicating stronger evidence in support of normality. The original formulation required use of the tables of the critical values of $W$ \citep{Henderson2006} at the most common levels of $\alpha$ and was limited to smaller samples with $n \in [3, 20]$ elements because the values of $\mathbf{b}$ and $\mathbf{V}^{-1}$ were known only for small samples at the time \citep{ComparisonThesis2010}. \cite{Royston1982} extended the upper limit of $n$  to $2000$ and presented a normalizing transformation algorithm suitable for computer implementation. The upper limit was further improved by \cite{Royston1995} who gave an algorithm AS 194 which allowed the test to be used for the samples with $n \in [3, 5000]$.
	
	\subsection{The Lilliefors Test}\label{subsec:liliefors}
	
	This test was introduced independently by \cite{Lilliefors1967} and \cite{vanSoest1967}. The LF test checks if the given sample $\mathbf{x}$ comes from a normal distribution whose parameters $\mu$ and $\sigma$ are taken to be the sample mean ($\overline{x}$) and standard deviation ($\text{sd}(\mathbf{x})$). It is equivalent to the Kolmogorov-Smirnov test of goodness-of-fit \citep{Kolmogorov1933} for those particular choices of $\mu$ and $\sigma$ \citep{Wijekularathna2019}. The LF test is conducted as follows. If the sample at hand comes from $N(\overline{x},\text{sd}(\mathbf{x})^2)$, then its transformation $\mathbf{z}=[z_1,z_2,\ldots,z_n]$, where:
	\begin{equation}
	z_i = \frac{x_i-\overline{x}}{\text{sd}(\mathbf{x})}
	\end{equation}
	should follow $N(0, 1)$. The difference between the EDF of $\mathbf{z}$, $edf_{\mathbf{z}}$, and the CDF of $N(0, 1)$, $\Phi$, quantifies how well the sample $\mathbf{x}$ fits the normal distribution.
	In the LF test, that difference is calculated as follows 
	\begin{equation}\label{eq:lf-statistic}
	D = \max\limits_{i=1}^{n}\left|edf_{\mathbf{z}}(z_i)-\Phi(z_i)\right|
	\end{equation}
	Higher values of $D$ indicate greater deviation from normality.
	
	\subsection{The Anderson-Darling Test}\label{subsec:anderson-darling}
	Whereas the Lilliefors test focuses on the largest difference between the sample's EDF and the CDF of the hypothesized model distribution, the AD test calculates the expected weighted difference between those two \citep{AndersonDarling1954}, with the weighting function designed to make use of the specific properties of the model. 	The AD statistic is:
	\begin{equation}\label{eq:ad-statistic_definition}
	A= n\int_{-\infty}^{+\infty}\left[F^*(x)-edf(x)\right]^2\psi(F^*(x))dF^*(x)
	\end{equation}
	where $\psi$ is the weighting function, $edf$ is the sample's EDF, and $F^*$ is the CDF of the distribution we want to test. When testing for normality, the weighting function is chosen to be sensitive to the tails \citep{AndersonDarling1954,Wijekularathna2019}:
	\begin{equation}
	\psi(t)=\left[t(1-t)\right]^{-1}
	\end{equation}
	Then, the statistic \eqref{eq:ad-statistic_definition} can be calculated in the following way \citep{Thode2002testing,nortest2015}:
	\begin{equation}\label{eq:ad-statistic}
	A = -n -\frac{1}{n}\sum_{i=1}^{n}\left[2i-1\right]\left[\ln\Phi(z_{(i)}) + \ln\left(1-\Phi(z_{(n-i+1)})\right)\right]
	\end{equation}
	where $[z_{(1)},z_{(2)},\ldots,z_{(n)}]$ ($z_{(i)} \leq z_{(i+1)}, i=1,2,\ldots,n-1$) is the ordered permutation of $\mathbf{z}=[z_1,z_2,\ldots,z_n]$ that is obtained from $\mathbf{x}$ as in the LF test. The $p$-values are computed from the modified statistic \citep{nortest2015,Stephens1986}: 
	\begin{equation}
	A(1+0.75/n+2.25/n^2)
	\end{equation}
	As noted by \cite{Wijekularathna2019}, the AD test is a generalization of the Cramer - von Mises (CVM) test \citep{Cramer1928,vonMises1928}. When $\psi(\cdot)=1$, the AD test's statistic reduces to that of the CVM test:
	\begin{equation}\label{eq:cvm-statistic_definition}
	C= n\int_{-\infty}^{+\infty}\left[F^*(x)-edf(x)\right]^2dF^*(x)
	\end{equation}
	Because $\psi(\cdot)$ takes into account the specific properties of the model distribution, the AD test may be more sensitive than the CVM test \citep{Wijekularathna2019}. In both AD and CVM tests, larger values of the statistics indicate stronger arguments in favor of non-normality.
	
	\subsection{The Jarque-Bera Test}
	The Jarque--Bera test \citep{JarqueBera1980,Jarque1987}, checks how much the sample's skewness ($\sqrt{\beta_1}$) and kurtosis ($\beta_2$) match those of normal distributions. Namely, for each normal distribution it holds that $\sqrt{\beta_1}=0$ and $\beta_2=3$. The statistic of the test, as originally defined by \cite{JarqueBera1980}, is computed as follows:
	\begin{equation}\label{eq:jb-statistic}
	J = n\left(\frac{\left(\sqrt{\beta_1}\right)^2}{6} + \frac{(\beta_2-3)^2}{24}\right) = \frac{\left(\sqrt{\beta_1}\right)^2}{6/n} + \frac{(\beta_2-3)^2}{24/n}
	\end{equation}
	
	We see that higher values of $J$ indicate greater deviation from the skewness and kurtosis of normal distributions. The same idea was examined by \cite{Bowman1975}. The asymptotic expected values of the estimators of skewness and kurtosis are $0$ and $3$, while the asymptotic variances are $6/n$ and $24/n$ for the sample of size $n$ \citep{Bowman1975}. The $J$ statistic is then a sum of two asymptotically independent standardized normals. However, the  estimator of kurtosis slowly converges to normality, which is why the original statistic was not useful for small and medium-sized samples \citep{Urzua1996}. \cite{Urzua1996} adjusted the statistic by using the exact expressions for the means and variances of the estimators of skewness and kurtosis:
	\begin{equation}\label{eq:ajb-statistic}
	J = n\left(\frac{\left(\sqrt{\beta_1}\right)^2}{c_1} + \frac{(\beta_2-c_2)^2}{c_3}\right)
	\end{equation}
	where:
	\begin{eqnarray*}
		c_1 & = & \frac{6(n-2)}{(n+1)(n+3)}\\
		c_2 & = & \frac{3(n-1)}{n+1}\\
		c_3 & = & \frac{24n(n-2)(n-3)}{(n+1)^2(n+3)(n+5)}
	\end{eqnarray*}
	which allowed the test to be applied to smaller samples.
	
	\subsection{The Kernel Test of Normality}
	As mentioned in Section \ref{sec:literature_review}, there are several kernel tests of goodness-of-fit. Just as our approach, they also represent a blend of machine learning and statistics. We evaluate the test of \cite{Wittawat2017} against our network in this study because that test is computationally less complex than the original kernel tests of goodness-of-fit proposed by \cite{Chwialkowski2016}, but comparable with them in terms of statistical power. Of the other candidates, the approach of \cite{Kellner2019} is of quadratic complexity and requires bootstrap, and a drawback of the approach of \cite{Lloyd2015}, as \cite{Wittawat2017} point out, is that it requires a model to be fit and new data to be simulated from it. Also, this approach fails to exploit our prior knowledge on the characteristics of the	distribution for which goodness-of-fit is being determined. So, the test formulated by \cite{Wittawat2017} was our choice as the representative of the kernel tests of goodness-of-fit. It is a distribution-free test, so we first describe its general version before we show how we used it to test for normality. The test is defined for multidimensional distributions, but we present it for the case of one-dimensional distributions for simplicity and because we are interested in one-dimensional Gaussians.
	
	Let $\mathcal{F}$ be the RKHS of real-valued functions over $\mathcal{X}\in\R$ with the reproducing kernel $k$. Let $q$ be the density of model $\Psi$. As in \cite{Chwialkowski2016}, a Stein operator $T_{q}$ \citep{Stein1972} can be defined over $\mathcal{F}$:
	\begin{equation}
	(T_{q}f)(x) = \frac{\partial\log q(x)}{\partial x}f(x)+\frac{\partial f(x)}{\partial x}
	\end{equation}
	Let us note that for
	\begin{equation}
	\xi_q(x, \cdot)=\frac{\partial\log q(x)}{\partial x}k(x, \cdot) + \frac{\partial k(x, \cdot)}{\partial x}
	\end{equation}
	it holds that:
	\begin{equation}
	(T_q f)(x) = \inner{f}{\xi_q(x, \cdot)}_{\mathcal{F}}
	\end{equation}
	
	If $Z \sim \Psi$, then $\Exp(T_{q}f)(Z)= 0$ \citep{Chwialkowski2016}.
	Let $X$ be the random variable which follows the distribution from which the sample $\mathbf{x}$ was drawn. The Stein discrepancy $S_q$ between $X$ and $Z$ is defined as follows \citep{Chwialkowski2016}:
	\begin{equation}
	\begin{split}
	S_{q}(X) &= \sup_{\norm{f} < 1} \left[ \Exp (T_{q}f)(X)-\Exp (T_{q}f)(Z) \right]= \sup_{\norm{f}<1}\Exp (T_{q}f)(X)\\ &= \sup_{\norm{f} < 1}\inner{f}{\Exp\xi_q(X,\cdot)}_{\mathcal{F^d}} = \norm{g(\cdot)}_{\mathcal{F}}
	\end{split}
	\end{equation}
	where $g(\cdot)=\Exp\xi_q(X,\cdot)$ is called the Stein witness function and belongs to $\mathcal{F}$.  \cite{Chwialkowski2016} show that if $k$ is a cc-universal kernel \citep{Carmeli2010}, then $S_{q}(X)=0$ if and only if $X \sim \Psi$, provided a couple of mathematical conditions are satisfied.
	
	\cite{Wittawat2017} follow the approach of \cite{Chwialkowski2015} for kernel two-sample tests and present the statistic that is comparable to the original one of \cite{Chwialkowski2016} in terms of power, but faster to compute.  The idea is to use a real analytic kernel $k$ that makes the witness function $g$ real analytic. In that case, the values of $g(v_1),g(v_2),\ldots,g(v_m)$ for a sample of points $\{v_j\}_{j=1}^{m}$, drawn from $X$,  are almost surely zero w.r.t. the density of $X$ if $X\sim\Psi$. \cite{Wittawat2017} define the following statistic which they call the finite set Stein discrepancy:
	\begin{equation}
	FSSD^2 = \frac{1}{m}\sum_{j=1}^{m}g(v_i)^2
	\end{equation}
	If $X\sim\Psi$, $FSSD^2=0$ almost surely.  \cite{Wittawat2017} use the following estimate for $FSSD^2$:
	\begin{equation}\label{eq:fssd_2_estimate}
	FSSD^2 = \frac{2}{n(n-1)}\sum_{i=2}^{n}\sum_{j < i}\Delta(x_i,x_j)
	\end{equation}
	where 
	\begin{equation}
	\Delta(x_i, x_j) = \frac{1}{m}\sum_{\ell=1}^{m}\xi_q(x_i,v_{\ell})\xi_q(x_j,v_{\ell})
	\end{equation}
	
	In our case, we want to test if $\Psi$ is equal to any normal distribution. Similarly to the LF test, we can use the sample estimates of the mean and variance as the parameters of the normal model. Then, we can randomly draw $m$ numbers from $N(\overline{x}, \text{sd}(\mathbf{x})^2)$ and use them as points $\{v_j\}_{j=1}^{m}$, calculating the estimate \eqref{eq:fssd_2_estimate} for $\mathbf{x}$ and $N(\overline{x}, \text{sd}(\mathbf{x}))$. For kernel, we chose the Gaussian kernel as it fulfills the conditions laid out by \cite{Wittawat2017}. To set its bandwidth, we used the median heuristic \citep{Garreau2017}, which sets it to the median of the absolute differences $|x_i-x_j|$ ($x_i, x_j \in \mathbf{x}, 1\leq i < j \leq n$). The exact number of locations, $m$, was set to $10$.
	
	Since $g$ is always zero if the sample comes from the normal distribution, the larger the value of $FSSD^2$, the more likely it is that the sample came from a non-normal distribution. We will refer to this test as the FSSD test from now on.
	
	\subsection{The Statistic-Based Neural Network}
	Since the neural networks are designed with a fixed-size input in mind, and samples can have any number of elements, \cite{Sigut2006} represent the samples with the statistics of several normality tests which were chosen in advance. The rationale behind this method is that, taken together, the statistics of different normality tests examine samples from  complementary perspectives, so a neural network that combines the statistics could be more accurate than individual tests. \cite{Sigut2006} use the estimates of the following statistics:
	\begin{enumerate}
		\item skewness:
		\begin{equation}
			\sqrt{\beta_1} = \frac{m_3}{m_2^{3/2}}
		\end{equation}
		where $m_u=\frac{1}{n}\sum_{i=1}^{n}(x_i-\overline{x})^u$,
		\item kurtosis:
		\begin{equation}
			\sqrt{\beta_2} = \frac{m_4}{m_2^2}
		\end{equation}
		\item the $W$ statistic of the Shapiro-Wilk test (see Equation \eqref{eq:w-statistic}),
		\item the statistic of the test proposed by \cite{Lin1980}:
		\begin{equation}
		\begin{split}
			Z_p &= \frac{1}{2}\ln\frac{1+r}{1-r} \\
			r &= \frac{\sum_{i=1}^{n}(x_i-\overline{x})(h_i-\overline{h})}{\sqrt{\left(\sum_{i=1}^{n}(x_i-\overline{x})^2\right)\left(\sum_{i=1}^{n}(h_i-\overline{h})^2\right)}}\\
			h_i &= \left(\frac{\sum_{j\neq i}^{n}x_j^2-\frac{1}{n-1}\left(\sum_{j\neq i}^{n}x_j\right)^2}{n}\right)^{\frac{1}{3}}\\
			\overline{h} &= \frac{1}{n}h_i
		\end{split}
		\end{equation}
		\item and the statistic of the Vasicek test \citep{Vasicek1976}:
		\begin{equation}
			K_{m,n} = \frac{n}{2m\times\text{sd}(\mathbf{x})}\left(\prod_{i=1}^{n}(x_{(i+m)}-x_{(i-m)})\right)^{\frac{1}{n}}
		\end{equation}
		where $m$ is a positive integer smaller than $n/2$, $[x_{(1)}, x_{(2)}, \ldots, x_{(n)}]$ is the non-decreasingly sorted sample $\mathbf{x}$, $x_{(i)}=x_{(1)}$ for $i<1$, and $x_{(i)}=x_{(n)}$ for $i>n$. 
	\end{enumerate}
	
It is not clear which activation function \cite{Sigut2006} use. They trained three networks with a single hidden layer containing $3$, $5$ and $10$ neurons, respectively. One of the networks was designed to take the sample size into account as well so that it can be more flexible. Just as the other two, the network showed that it was capable of modeling posterior Bayesian probabilities of the input samples being normal. \cite{Sigut2006} focus on the samples with no more than $200$ elements.

In addition to our network, presented in Section \ref{sec:neural_networks}, we trained one that follows the philosophy of \cite{Sigut2006}. We refer to that network as Statistic-Based Neural Network (SBNN) because it expects an array of statistics as its input. More precisely, prior to being fed to the network, each sample $\mathbf{x}$ is transformed to the following array:
\begin{equation}
	[\sqrt{\beta_1}, \beta_2, W, Z_p, K_{3,n}, K_{5, n}, n]
\end{equation}
just as in \cite{Sigut2006} ($n$ is the sample size). ReLU was used as the activation function. To make comparison fair, we trained SBNN in the same way as our network which we design in Section \ref{sec:neural_networks}.

	\section{Neural Networks for Identifying Normal Distributions}\label{sec:neural_networks}
	The problem of testing for normality, designed as a statistical hypothesis testing problem in Definition \ref{def:testing_for_normality}, can be straightforwardly cast as a binary classification problem.
	\begin{definition}\label{def:normality_as_classification}
		Given a sample $\mathbf{x}=[x_1, x_2, \ldots, x_n]$ drawn from a distribution modeled by a random variable $X$, classify $X$
		\begin{itemize}
			\item as following a normal distribution, by assigning it label $1$; or
			\item as following a non-normal distribution, by assigning it label $0$.
		\end{itemize}
	\end{definition}
	Here, class $1$ will denote the set of all normal distributions, and class $0$ will represent the set of all non-normal distributions. 
	
	The core idea of our approach is to create a dataset consisting of the normal and non-normal samples of different sizes, drawn from various normal and non-normal distributions, and train a neural network on that dataset as if this was any other binary classification problem. Moreover, since we can construct as large a dataset as we want, we are not limited by the data availability as in other learning problems. We can use as much data as our computational capacities allow. 
	
	However, since samples can contain any number of elements, and training a network for each imaginable sample size is impossible, a preprocessing step is required to transform the given sample, no matter how large, to a vector with fixed number of dimensions. We call those vectors descriptors and describe them in detail in Section \ref{subsec:descriptors}. 
	
	The networks we considered in this study are simple feedforward networks with two hidden layers at most. We used ReLU as the activation function. The final layer was designed to be the softmax layer. That means that for each sample $\mathbf{x}$, the output is a single value $p_1(\mathbf{x}) \in [0, 1]$ which may be interpreted as the probability that $\mathbf{x}$ belongs to class $1$, i.e. that it comes from a normal distribution.  In fact, minimizing the log-loss (as we did) ensures that the network's output $p_1(\mathbf{x})$ is an estimate of the Bayesian posterior probability that $\mathbf{x}$ has been drawn from a normal distribution \citep{Hampshire1991,Richard1991}. The label assigned to $\mathbf{x}$, $\hat{y}(\mathbf{x})$, is determined using the following decision rule:
	\begin{equation}
		\hat{y}(\mathbf{x}) = \begin{cases}
		1, & p_1(\mathbf{x}) \geq 0.5\\
		0, & p_1(\mathbf{x}) < 0.5
		\end{cases}
	\end{equation}
	
	Even though the design described above is not complex, it turned out that it was able to successfully learn efficient classification rules from the data.
	
	\subsection{Descriptors}\label{subsec:descriptors}
	Neural networks require that all the objects they classify have the same dimensions: this means fixed number and order of features. On the other hand, samples of data, analyzed in everyday practice of statistics, can have any number of elements. Even though it is the most natural way to represent samples by their actual content, i.e. by arrays containing the elements of the samples, such an approach is not convenient for machine-learning algorithms as it requires training a separate classifier for each possible value of the sample size, $n$, which is not a feasible strategy.
	
	\cite{Sigut2006} approach the problem by representing samples with the statistics of several normality tests, just as \cite{Wilson1990}. We propose to extend their approach in the following way. First of all, we note that samples can be represented not just by the statistics of normality tests, but also by a number of descriptive statistics such as the mean, median, standard deviation, etc. Then, we use the same idea as the LF and kernel tests: if a random variable $X$ follows any normal distribution $N(\mu, \sigma^2)$, then it can be transformed to follow the standard normal $N(0,1)$, as formally stated by the following theorem.
	\begin{theorem}[\cite{Grinstead2003}]\label{the:standardization}
		If $X \sim N(\mu, \sigma^2)$, then:
		\begin{equation}
		Z = \frac{X - \mu}{\sigma} \sim N(0,1)
		\end{equation}
	\end{theorem}
	If the sample under consideration comes from a normal distribution, then its standardization should produce a sample that more or less looks like as if it was drawn from $N(0,1)$. A structure with which we choose to represent samples should make use of Theorem \ref{the:standardization}. Also, it should contain not only a number of summarizing statistics, but some original elements from raw samples as well. Therefore, we propose to consider evenly spaced quantiles of standardized samples, as they should be able to capture the given sample's likeness to $N(0,1)$ as well as expose the network to the selected elements from the raw sample. The quantiles combined with the mean, standard deviation, minimum, maximum, and possibly other statistics, could yield good classification results and enable us to train networks capable of handling samples with any number of elements. In addition to statistics, there is one more piece of information that we believe is important and should be contained in a sample representation, and that is the sample size. We hypothesize that if a network takes the sample size into consideration, then it will be able to learn different classification rules for different sample sizes. We call such structures descriptors and formally introduce them in the following definition.

	\begin{definition}\label{def:descriptor}
		Let $\mathbf{x}=[x_1,x_2,\ldots,x_n]$ be a sample with $n$ elements. The descriptor of $\mathbf{x}$ is the array
		\begin{equation}\label{eq:descriptor}
		[h_{q}(\mathbf{z}), h_{2q}(\mathbf{z}), \ldots, h_1(\mathbf{z}), n, \overline{x}, \text{\emph{sd}}(\mathbf{x}), \min(\mathbf{x}), \max(\mathbf{x}), \text{\emph{med}}(\mathbf{x})]
		\end{equation}
		where $\mathbf{z} = [z_1, z_2, \ldots, z_n], z_i = (x_i-\overline{x})/\text{\emph{sd}}(\mathbf{x})$ for  $i=1,2,\ldots, n$,  $h_{p}(\mathbf{z}) = \min\{z_i : F(z_i) \geq p\}$ for $p \in (0, 1]$, $q \in(0, 1]$ and $F$ is the EDF of $\mathbf{z}$. The mean and standard deviation of $\mathbf{x}$ are $\overline{x}$ and $\text{\emph{sd}}(\mathbf{x})$, its minimal and maximal values are  $\min(\mathbf{x})$ and $\max(\mathbf{x})$, and  $\text{\emph{med}}(\mathbf{x})$ represents its median.
	\end{definition}
	Transforming samples into descriptors allows us to train a single classifier once and use it to test normality of a sample with any number of elements.
	
	The value of $q$, that determines which quantiles from a sample get included in its descriptor, should be set to the number that will not make the descriptors too large, so that training is not slow, keeping them representative of the samples at the same time as well. We determined the value of $q$ via cross-validation.
	
	We name the networks that use descriptors the descriptor-based neural networks (DBNN), which is how we will refer to them in the remainder of the paper. 
	
	\section{Datasets}\label{sec:datasets}
	Since no public dataset of samples drawn from various distributions and suitable for training was available at the time of our research, we have created $6$ different datasets to construct and evaluate our network. Four of them are artificial, and two are based on the real-world data.
	
	We name the artificial datasets $\mathcal{A}$-$\mathcal{D}$ and describe them as follows.

	Set $\mathcal{A}$ consists of $13050$ normal and as much non-normal samples. The samples are of the sizes $n=10, 20, \ldots, 100$. The set is balanced by labels and sample sizes. The normal samples were drawn from randomly selected normal distributions. Each distribution was specified by setting its mean to a random number from $[-100, 100]$ and deviation to a random number from $[1, 20]$. The non-normal samples were simulated from the Pearson family of distributions. \cite{Pearson1895,Pearson1901,Pearson1916} identified twelve different types of distributions. The density $f$ of each distribution in this family is obtained by solving the following differential equation \citep{Karvanen2000}:
		\begin{equation}\label{eq:pearson_equation}
		\frac{f'(x)}{f(x)} = \frac{x-a}{b_0+b_1x+b_2x^2}
		\end{equation}
		where $a, b_0, b_1$ and $b_2$ are the distributions's parameters. They depend on the central moments as follows
		\begin{align}
		b_1 & = a = - \frac{m_3(m_4+3m_2^2)}{C}\\
		b_0 & = - \frac{m_2(4m_2m_4 - 3m_3^2)}{C}\\
		b_2 & = - \frac{2m_2m_4-3m_3^2-6m_2^3}{C}
		\end{align}
		where $C=10m_4m_2-12m_3^2-18m_2^3$ and $m_j$ is the $j$-th central moment of the distribution. Given $\mu$, $\sigma$, $\sqrt{\beta_1}$ and $\beta_2$ ($\sqrt{\beta_1}=m_3/m_2^{3/2}$ is skewness, $\beta_2=m_4/m_2^2$ is kurtosis), it is possible to calculate $a$, $b_0$, $b_1$, and $b_2$, and simulate samples from the distribution they specify. 
		
		When generating non-normal samples, for each feasible combination of $\sqrt{\beta_1} \in [-30, -29.5, \ldots, 29.5, 30]$ and $\beta_2 \in [0.0, 0.5, \ldots, 39.5, 40 ]$ we randomly chose $\mu \in [-100,100]$ and $\sigma \in [1, 20]$. We avoided the combination of $\sqrt{\beta_1}$ and $\beta_2$ characteristic to normal distributions. More precisely, we did not consider  non-normal distributions for which the following held:  $\sqrt{\beta_1}=0$ and $\beta_2=3$. A lot of non-normal distributions was covered in this way. 
		\begin{itemize}
			\item We used $70\%$ of $\mathcal{A}$ for cross-validation. We will denote that subset as $\mathcal{A}_{cv}$. Just as $\mathcal{A}$, $\mathcal{A}_{cv}$ is also balanced. 
			\item  The rest $30\%$ of $\mathcal{A}$ was reserved for initial testing. We will refer to that subset as $\mathcal{A}_{test}$. It is balanced too.
		\end{itemize}
	
	Set $\mathcal{B}$ was generated in the same way as $\mathcal{A}$. The only difference is that $\mathcal{B}$ contains the samples of the sizes $n=5, 15, \ldots, 95$. The samples of those sizes were not used to train the network. The purpose of $\mathcal{B}$ was to check the net's robustness with respect to the sample size.
	
	Set $\mathcal{C}$ contains the samples from the four groups $G_1-G_4$ of non-normal distributions that were used by \cite{Esteban2001} and \cite{Noughabi2011} to empirically evaluate several standard tests of normality. Those groups are presented in Table  \ref{tab:alternative_distributions}. For each sample size $n=10,20,30,\ldots,100$ and $i=1,2,3,4$, we generated $10000$ samples with $n$ elements from the distributions in group $G_i$, iterating over them cyclicly and drawing a sample of $n$ elements until we drew $10000$ such samples. We used this set to estimate more precisely the ability of our network to detect non-normal distributions.
	
	Set $\mathcal{D}$ was created in the same was as $\mathcal{A}$ and used to additionally evaluate and analyze the DBNN.
	
	\begin{table}
		\centering
		\caption{Specification of the non-normal distributions in dataset $\mathcal{C}$}
		\label{tab:alternative_distributions}
		\begin{tabular}{llll}
			\noalign{\smallskip}\hline\noalign{\smallskip}
			Group &  Support & Shape & Distributions\\
			\noalign{\smallskip}\hline\noalign{\smallskip}
			$G_1$ & $(-\infty, \infty)$ & Symmetric & $t(1)$ \\
			\multicolumn{3}{l}{ } & $t(3)$  \\
			\multicolumn{3}{l}{ } & standard logistic \\
			\multicolumn{3}{l}{ } & standard Laplace \\
			\midrule
			$G_2$ & $(-\infty, \infty)$ & Asymmetric & Gumbel($0, 1$) \\
			\multicolumn{3}{l}{ } & Gumbel($0, 2$) \\
			\multicolumn{3}{l}{ } & Gumbel($0, 1/2$) \\
			\midrule
			$G_3$ & $(0,\infty)$ & Various  & Exponential with mean $1$ \\
			\multicolumn{3}{l}{ } & Gamma($1, 2$) \\
			\multicolumn{3}{l}{ } & Gamma($1, 1/2$) \\
			\multicolumn{3}{l}{ } & Lognormal($0, 1$) \\
			\multicolumn{3}{l}{ } & Lognormal($0, 2$) \\
			\multicolumn{3}{l}{ } & Lognormal($0, 1/2$) \\
			\multicolumn{3}{l}{ } & Weibull($1, 1/2$) \\
			\multicolumn{3}{l}{ } & Weibull($1, 2$) \\
			\midrule
			$G_4$ & $(0, 1)$ & Various & Uniform($0, 1$) \\
			\multicolumn{3}{l}{ } & Beta($2$, $2$) \\
			\multicolumn{3}{l}{ } & Beta($0.5$, $0.5$)\\
			\multicolumn{3}{l}{ } & Beta($3$, $1.5$)\\
			\multicolumn{3}{l}{ } & Beta($2$, $1$)\\
			\noalign{\smallskip}\hline
		\end{tabular}
	\end{table}
	
	As sets $\mathcal{A}$-$\mathcal{D}$ are artificially created, we evaluated our network on two real-world datasets to gain a better insight into its practical usefulness.
	
\begin{figure}
		\centering
		\caption{The EDF of the heights of women in the age group $[24, 33)$ and EDFs of simulated normal samples drawn from the normal distribution with the mean and variance set to the original sample estimates }\label{fig:cdfs_heights}
		\includegraphics[width=0.7\textwidth]{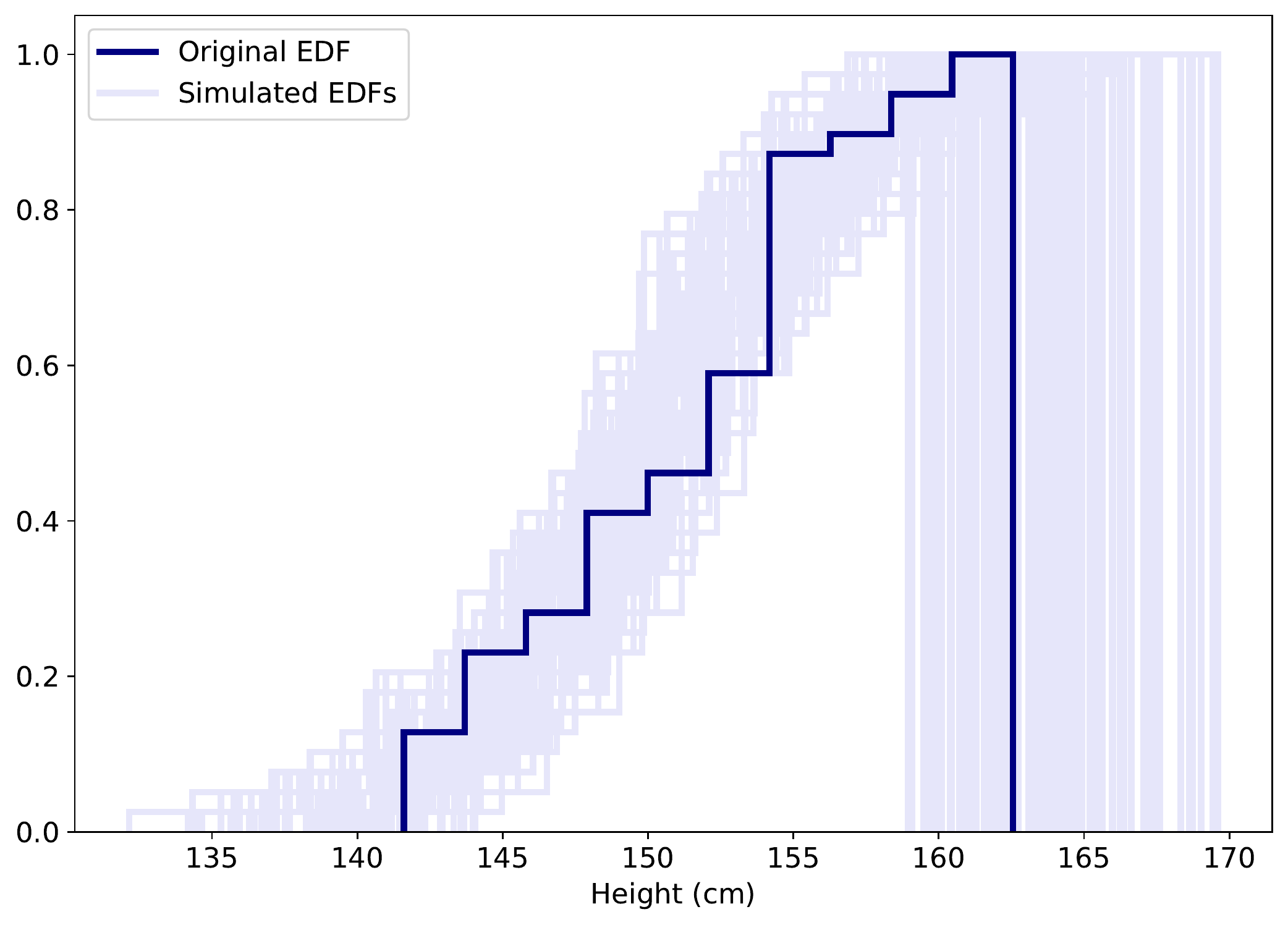}
	\end{figure}

	The first real-life set, $\mathcal{R}_{height}$, is based on a part\footnote{https://github.com/rmcelreath/rethinking/blob/master/data/Howell1.csv} of data on the heights of the Dobe !Kung people collected by \cite{Howell2010,Howell2017}. The part contains the heights (in centimeters) of $545$ people, including the information on sex and age. For each age span $[18, 27), [19, 28), \ldots, [80, 89)$, we sampled the heights of men and women separately, thus creating $124$ samples representing different groups of people. To confirm that the heights in each group are normally distributed, for each sample $\mathbf{x}$ we drew $100$ random same-sized samples from $N(\overline{x}, \text{sd}(\mathbf{x})^2)$ and plotted their EDFs. An example is presented in Figure \ref{fig:cdfs_heights} for women in the age group $[24, 33)$. We see that the real sample's EDF appears to be same as those of simulated normal samples, so we can safely assume that the distribution of the heights in the group is normal. The plots are similar for all the other groups, so we concluded the same about their distributions. The sample sizes vary from $3$ to $46$.

\begin{figure}
		\centering
		\caption{The empirical density of the earthquake magnitudes in the Northern California Earthquake Catalog and Phase Data }\label{fig:magnitude_distribution}
		\includegraphics[width=0.7\textwidth]{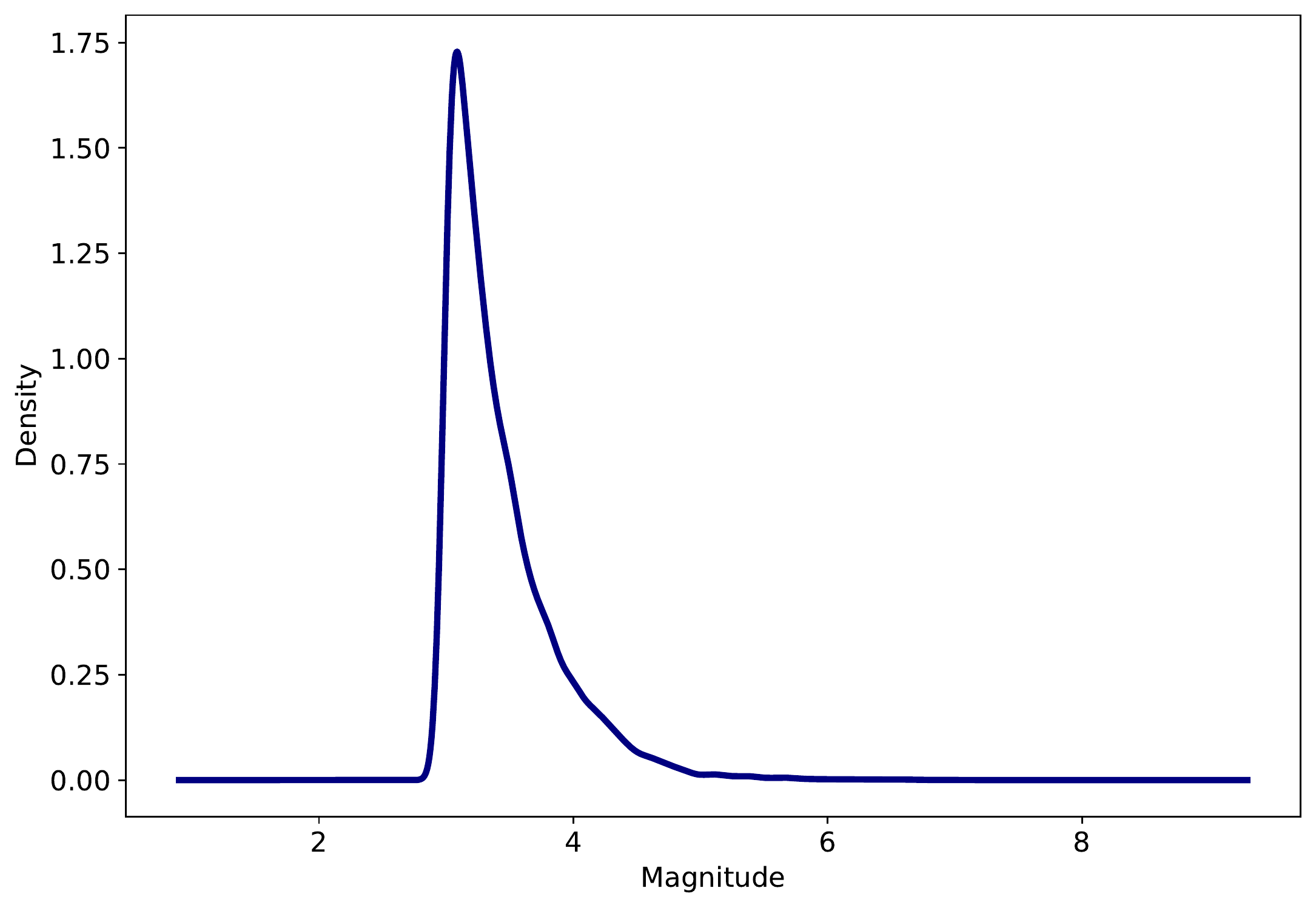}
	\end{figure}

	The second real-world set, $\mathcal{R}_{earthquake}$, was based on the Northern California Earthquake Catalog and Phase Data \citep{BDSN2014,NCEDC2014}. The catalog contains the records on $17811$ earthquakes that took place in California between July 1966 and December 2019 with the magnitudes of at least $3$ degrees. The distribution of the magnitudes in the catalog is presented in Figure \ref{fig:magnitude_distribution}. As we can see, it is not normal. For each $n=5,10,\ldots,95,100$, we created $1000$ samples by randomly drawing $n$ magnitudes from the catalog. 
	
	\section{Evaluation}\label{sec:evaluation}
	The cross-validation results are presented in Section \ref{subsec:cross-validation}. Using the Boruta algorithm \citep{Kursa2010}, we inspected the importance of the features included in the descriptors to make sure that they are well-defined. The results can be found in Section \ref{subsec:boruta}. We also checked if the constructed network is well-calibrated in the sense of \cite{Guo2017}. The details are in Section \ref{subsec:calibration}. The results of evaluation on sets $\mathcal{A}_{test}$ and $\mathcal{B}$ are presented in Sections \ref{subsec:results_A_test} and \ref{subsec:results_B}. The results of comparison with other methods for normality testing on sets $\mathcal{C}$ and $\mathcal{D}$ can be found in Section \ref{subsec:comparison}. Evaluation on real-world data is described in Section \ref{subsec:evaluation_on_real_world_data}. Finally, the results of DBNN's runtime analysis are reported in Section \ref{subsec:runtime_analysis}.
	
	\subsection{Cross-validation}\label{subsec:cross-validation}
	For training, we used ADAM \citep{Kingma2014} with early stopping, a positive regularization coefficient ($c$), and the maximal number of iterations set to $200$. The objective function to minimize was the log-loss.
	We conducted five-fold cross-validation with $10\%$ of $\mathcal{A}_{cv}$ reserved for validation after training. The hyperparameters considered were $q$, that determines the structure of the descriptors, the network's architecture (the number and sizes of the hidden layers), and $c$, the regularization coefficient. We conducted grid-search to set the hyperparameters.
	
	The results of cross-validation are presented in Table \ref{tab:cv_dbnn}. The best accuracy, $90.5\%$, was achieved by the network with two inner layers, the first with $100$ and the second with $10$ neurons, the parameter $q$ set to $0.1$, and trained with the regularization coefficient set to $0.1$. All the combinations of hyperparameters gave good results and were trained very quickly, with the whole cross-validation done in a couple of minutes.
	
	After cross-validation, we trained the network with the best combination of hyperparameters using the whole set $\mathcal{A}_{cv}$.
	
	\begin{table}
		\centering
		\caption{The results of cross-validation. The classification accuracy and training time are presented with the mean and standard deviation calculated over five validation steps. Architecture refers to the number and sizes of the hidden layers. The regularization coefficient is denoted as $c$, while $q$ represents the parameter controlling the structure of the descriptors.}
		\label{tab:cv_dbnn}
		\begin{tabular}{lllll}
			\toprule
			$q$ &  architecture & $c$ &  accuracy & time ($s$)\\
			\midrule
			$0.05$ &  $(100, 10)$ &  $0.1$ &  $0.904\pm0.007$ &  $3.207\pm1.162$ \\
			$0.05$ &  $(100, 10)$ &  $1$ &  $0.902\pm0.006$ &  $2.916\pm0.225$ \\
			$0.05$ &  $(100, 10)$ & $10$ &  $0.879\pm0.008$ &  $2.631\pm0.151$ \\
			$0.05$ &  $1000$ &  $0.1$ &  $0.903\pm0.010$ &  $7.419\pm2.162$ \\
			$0.05$ &  $1000$ &  $1$ &  $0.900\pm0.006$ &  $7.251\pm0.813$ \\
			$0.05$ &  $1000$& $10$ &  $0.876\pm0.006$ &  $5.479\pm1.180$ \\
			$0.10$ &  $(100, 10)$ &  $0.1$ &  $0.905\pm0.005$ &  $2.326\pm0.073$ \\
			$0.10$ &  $(100, 10)$ &  $1$ &  $0.905\pm0.006$ &  $2.433\pm0.163$ \\
			$0.10$ &  $(100, 10)$ & $10$ &  $0.876\pm0.008$ &  $1.939\pm0.092$ \\
			$0.10$ &  $1000$ &  $0.1$ &  $0.902\pm0.007$ &  $6.401\pm1.692$ \\
			$0.10$ &  $1000$ &  $1$ &  $0.900\pm0.006$ &  $5.929\pm1.033$ \\
			$0.10$ &  $1000$ & $10$ &  $0.866\pm0.009$ &  $4.235\pm0.529$ \\
			\bottomrule
		\end{tabular}
	\end{table}
	
	\subsection{Feature Importance Inspection}\label{subsec:boruta}
	We used the Boruta algorithm  formulated by \cite{Kursa2010} for feature selection with the aim to check if the dimensions of the descriptors are statistically correlated with membership in classes $1$ or $0$. 
	
	With $50$ iterations, maximal depth of trees set to $5$ levels, the quantile parameter $\tau=0.85$, significance threshold $\alpha=0.05$, and automatically constructed $113$ trees, Boruta found, using set $\mathcal{B}$, that all the features we included in the definition of the descriptors were statistically correlated with the labels.
	
	\subsection{Calibration Analysis}\label{subsec:calibration}
	The binary classifiers that output probabilities that the classified object belongs to the classes under consideration are called probabilistic. DBNN is such a classifier. Its  output $p_1(\mathbf{x})$ for input $\mathbf{x}$ can be interpreted as the posterior probability that the true class of $\mathbf{x}$, denoted as $y(\mathbf{x})$, is $1$, so we can obtain the probability for class $0$ by simple subtraction: $p_0(\mathbf{x}) = 1 - p_1(\mathbf{x})$.  
	We are interested in how reliable those probabilities are. For example, the probability $p_1(\mathbf{x})=0.9$ is reliable if  $90\%$ of the samples for which DBNN outputs $0.9$ do come from a normal distribution. Classifiers that output only reliable probabilities are called perfectly calibrated, and we formally introduce them in Definition \ref{def:perfect_calibration}.
	\begin{definition}[Perfect calibration \citep{Guo2017}]\label{def:perfect_calibration} A binary classifier is called perfectly calibrated if its posterior probability $p_1$ satisfies the following condition:
		\begin{equation}\label{eq:perfect_calibration}
		\Prob\left(Y=1\mid p_1(X)=t\right)=t,\quad \forall t\in[0,1]
		\end{equation}
	where $X$ is a random variable that models objects to be classified, and $Y$ is a random variable that models the true class of $X$.
	\end{definition}

	The reliability diagrams are often used to check if a classifier is calibrated \citep{Murphy1977,Murphy1987,Brocker2008}. The diagrams visualize the relationship between expected and empirical probabilities and are created as follows. First, the objects (here, samples) are sorted by the output probability of belonging to class $1$. Then, they are discretized in a number of same-sized bins. For each bin, the following values are calculated:
	\begin{itemize}
		\item the empirical proportion of the objects that belong to class $1$; 
		\item and the expected proportion of the objects that belong to class $1$, calculated with respect to the posterior probabilities as the mean posterior probability of the objects in the bin.
	\end{itemize}
	The reliability diagrams plot the empirical against expected proportions. If the classifier is perfectly calibrated, one can expect a $45\degree$ line. \citep{Fenlon2018}. 
	
\begin{figure}
		\centering
		\caption{Reliability diagrams of DBNN}\label{fig:reliability_diagrams}
		\subfloat[The reliability diagram made using the whole set $\mathcal{D}$ \label{subfig:reliability_diagram}]{\includegraphics[width=0.7\textwidth]{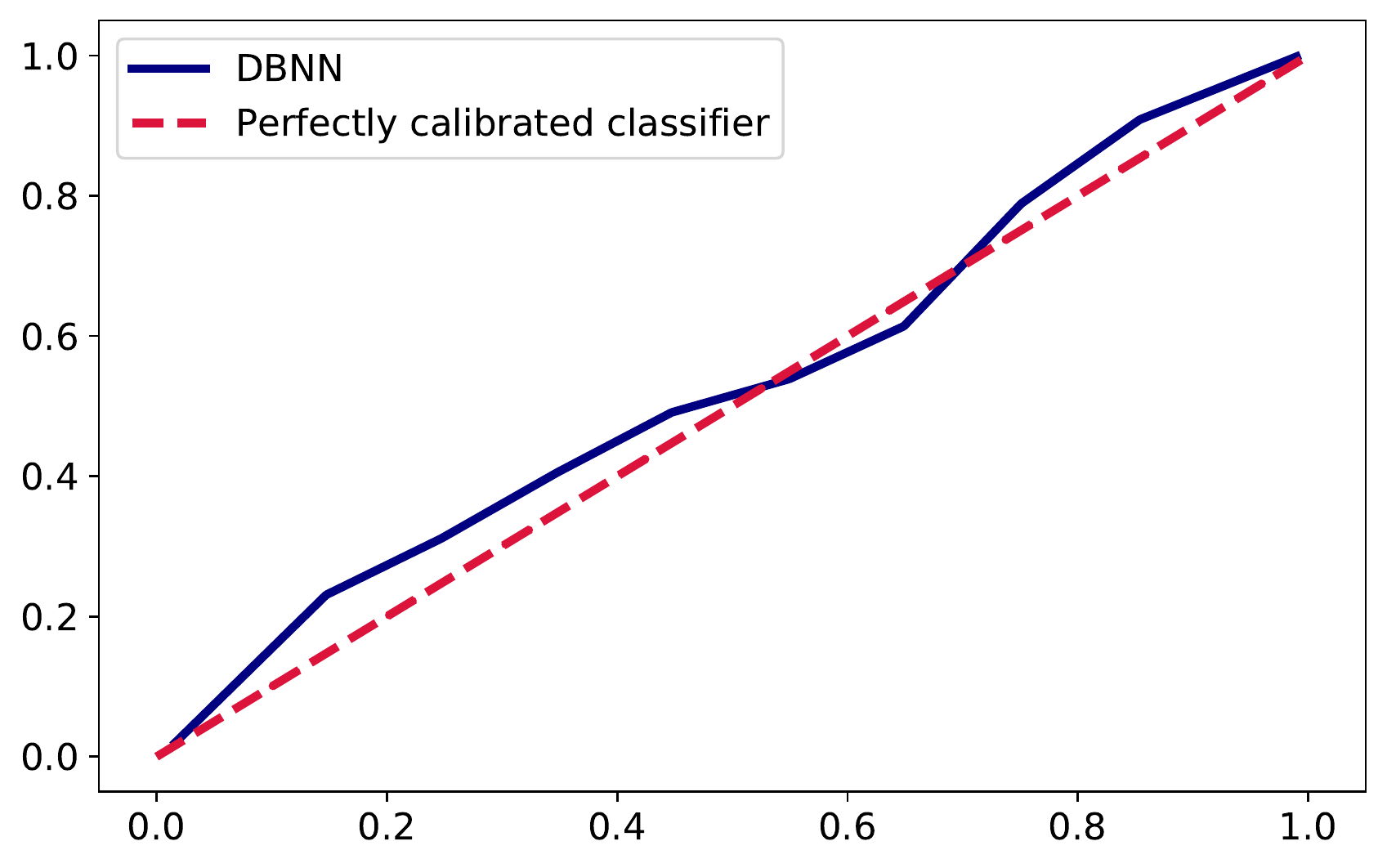}}\\
		\subfloat[The mean reliability graph and the graphs for the random subsets of  $\mathcal{D}$\label{subfig:reliability_diagrams_and_variability}]{\includegraphics[width=0.7\textwidth]{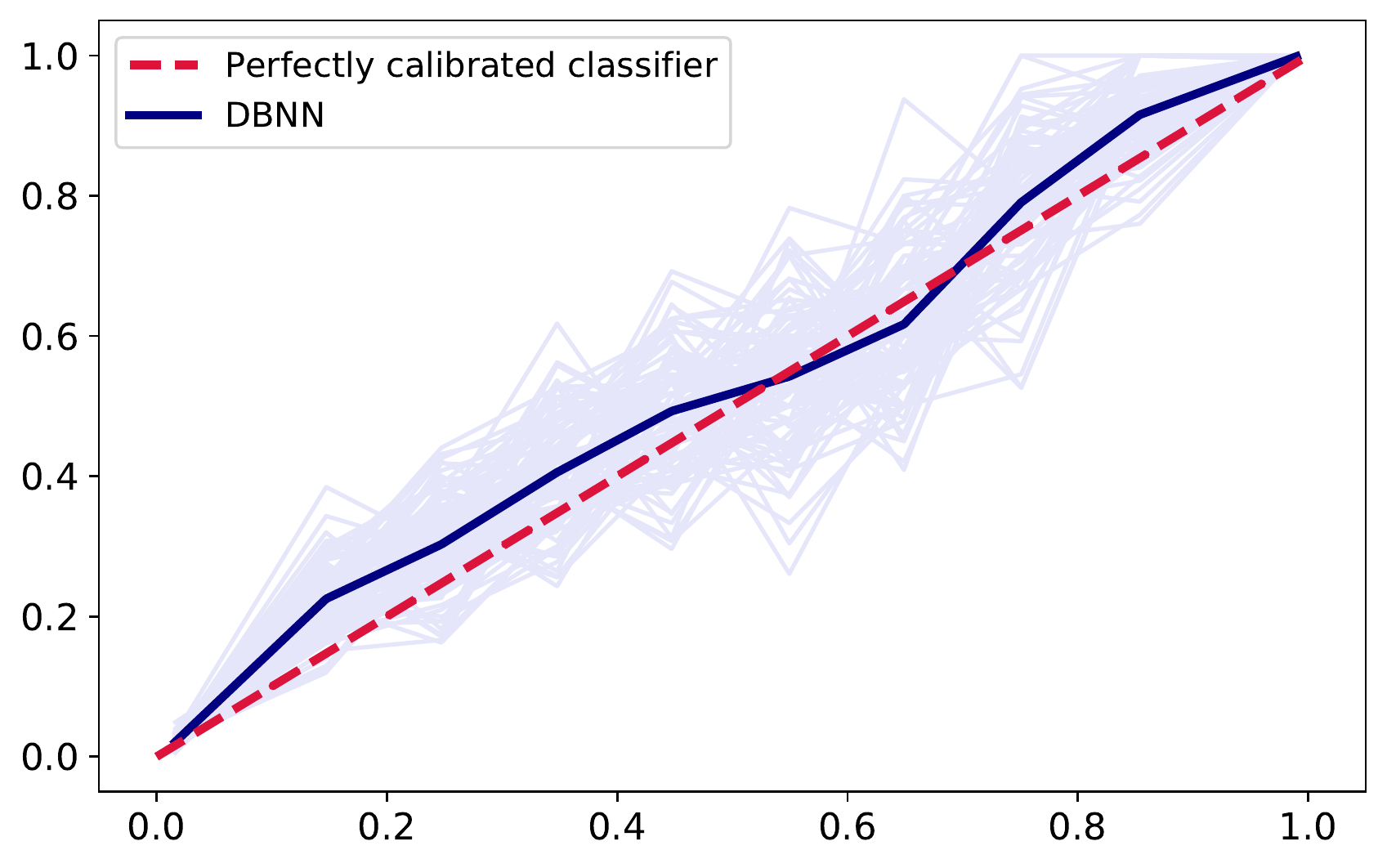}}
	\end{figure}

	We constructed the reliability diagram of DBNN using set $\mathcal{D}$. It is shown in Figure \subref*{subfig:reliability_diagram}. We see that it resembles the diagram of a perfectly calibrated classifier, but is not quite the same. In order to account for variability, we randomly selected $100$ subsets of $\mathcal{D}$, each containing $1000$ samples, and plotted their diagrams and the mean reliability graph in Figure \subref*{subfig:reliability_diagrams_and_variability}. We see that the reliability graphs constructed using the subsets capture the graph of the perfectly calibrated classifier. Therefore, we conclude that DBNN is a reliable classifier, but note that it is not perfectly calibrated.
	
	\subsection{Results on Set $\mathcal{A}_{test}$}\label{subsec:results_A_test}
\begin{figure}
		\centering
		\caption{Performance of DBNN on set $\mathcal{A}_{test}$: $TPR$, $PPV$, $TNR$ and $NPV$, per sample size $n$.}\label{fig:rezultati_A_test}
		\includegraphics[width=0.7\textwidth]{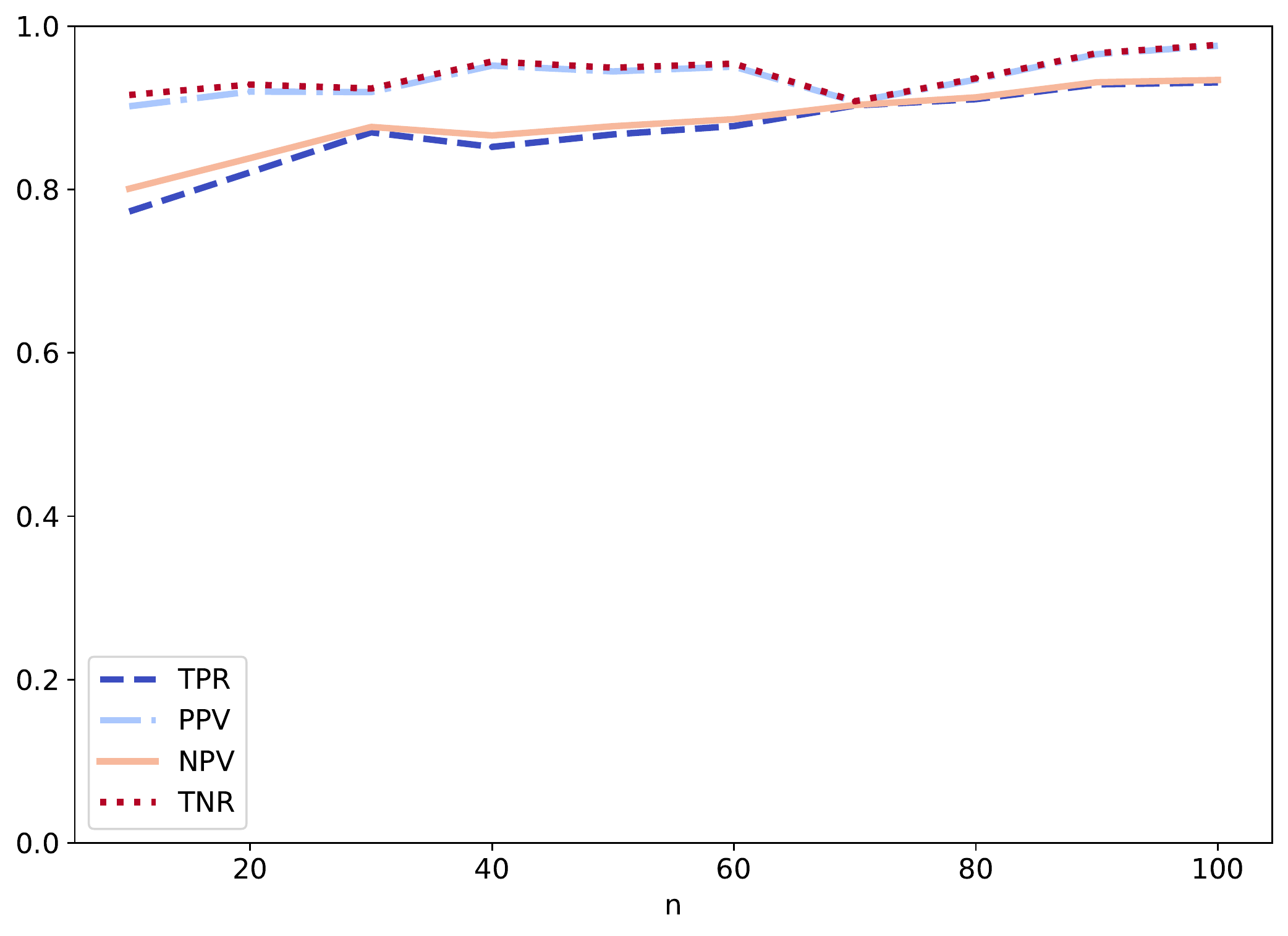}
	\end{figure}

	We calculated the following performance metrics:
	\begin{enumerate}
		\item Accuracy ($A$). 
		\item True positive rate ($TPR$), also known as recall. Its complement to $1$ is known as the Type I error rate in statistics.
		\item Positive predictive value ($PPV$), also known as precision: the proportion of normal samples among those classified as normal. 
		\item True negative rate ($TNR$). Known as power in statistical literature. Its complement to $1$ is called Type II error rate in statistics.
		\item Negative predictive value ($NPV$). Same as $PPV$, but for class $0$.
		\item The $F_1$ measure. The harmonic mean of $TPR$ and $PPV$.
		\item The area under the ROC curve ($AUROC$).
	\end{enumerate}
	The values of those metrics on $\mathcal{A}_{test}$ are presented in Table \ref{tab:results_A_test} and depicted in Figure \ref{fig:rezultati_A_test}. We see that DBNN achieved great accuracy and had the exceptional AUROC of almost $1$. All the scores consistently rise as the sample size $n$ increases.
	
	\begin{table}
		\centering
		\caption{Performance of DBNN on set $\mathcal{A}_{test}$}
		\label{tab:results_A_test}
		\begin{tabular}{llllllll}
			\toprule
			$n$   &    $A$ &     $TPR$ &     $PPV$ &     $TNR$ &     $NPV$ &      $F1$ &   $AUROC$ \\
			\midrule
			$10$      & $0.844$ & $0.773$ & $0.902$ & $0.916$ & $0.801$ & $0.832$ & $0.927$ \\
			$20$      & $0.875$ & $0.821$ & $0.920$ & $0.928$ & $0.838$ & $0.868$ & $0.957$ \\
			$30$      & $0.897$ & $0.870$ & $0.919$ & $0.923$ & $0.877$ & $0.894$ & $0.968$ \\
			$40$      & $0.904$ & $0.852$ & $0.952$ & $0.957$ & $0.866$ & $0.899$ & $0.971$ \\
			$50$      & $0.908$ & $0.867$ & $0.944$ & $0.949$ & $0.877$ & $0.904$ & $0.979$ \\
			$60$      & $0.916$ & $0.878$ & $0.950$ & $0.954$ & $0.886$ & $0.912$ & $0.984$ \\
			$70$      & $0.905$ & $0.903$ & $0.907$ & $0.908$ & $0.903$ & $0.905$ & $0.979$ \\
			$80$      & $0.923$ & $0.910$ & $0.934$ & $0.936$ & $0.913$ & $0.922$ & $0.986$ \\
			$90$      & $0.948$ & $0.929$ & $0.966$ & $0.967$ & $0.931$ & $0.947$ & $0.994$ \\
			$100$     & $0.954$ & $0.931$ & $0.976$ & $0.977$ & $0.934$ & $0.953$ & $0.994$ \\
			\midrule
			overall & $0.907$ & $0.873$ & $0.937$ & $0.941$ & $0.881$ & $0.904$ & $0.977$ \\
			\bottomrule
		\end{tabular}
	\end{table}
	
	\subsection{Results on Set $\mathcal{B}$}\label{subsec:results_B}
\begin{figure}
		\centering
		\caption{Performance of DBNN on set $\mathcal{B}$: $TPR$, $PPV$, $TNR$ and $NPV$, per sample size $n$.}\label{fig:results_B}
		\includegraphics[width=0.7\textwidth]{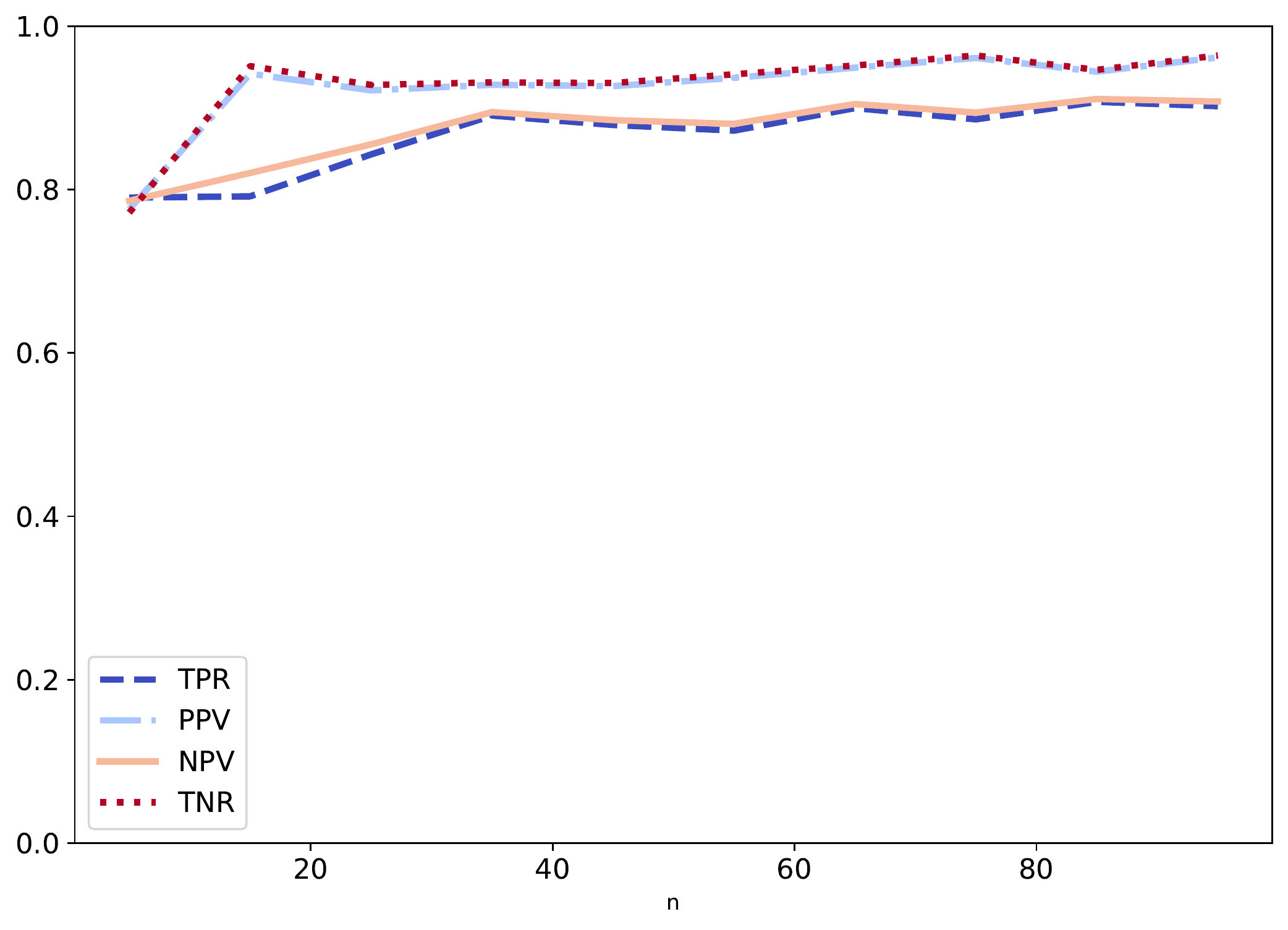}
	\end{figure}

\begin{figure}
		\centering
		\caption{Comparison of the DBNN's performance on sets $\mathcal{A}_{test}$ and $\mathcal{B}$}\label{fig:comparison_A_test_and_B}
		\includegraphics[width=0.7\textwidth]{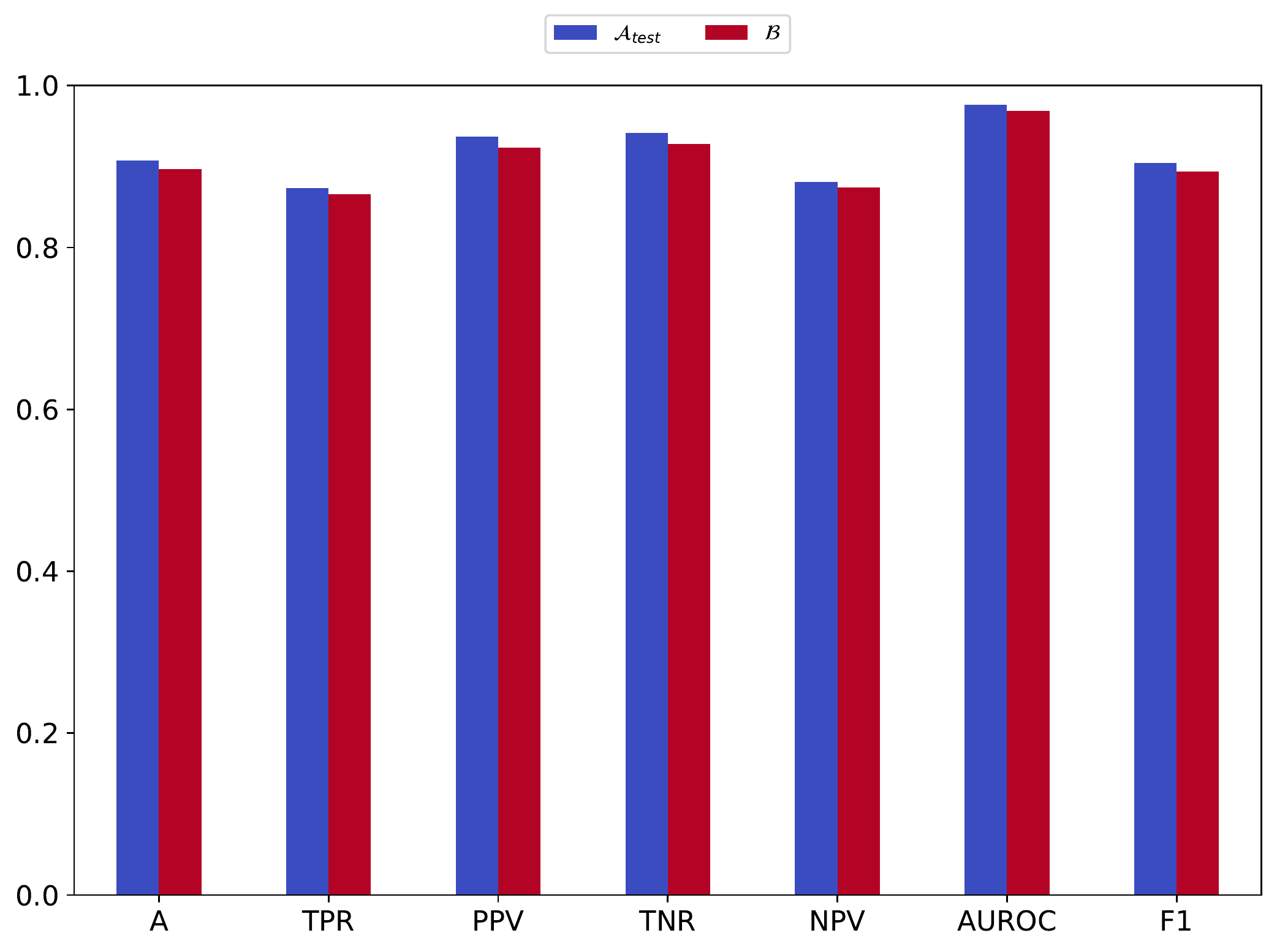}
	\end{figure}

	We calculated the same metrics on set $\mathcal{B}$ as for set $\mathcal{A}_{test}$ in order to see how DBNN handles the samples of sizes not seen during training. The results are presented in Table \ref{tab:results_B} and Figure \ref{fig:results_B}. The metrics on both sets are visualized in Figure \ref{fig:comparison_A_test_and_B}.
	
	The performance of DBNN on set $\mathcal{B}$ is comparable to its performance on set $\mathcal{A}_{test}$.
	
	\begin{table}
		\centering
		\caption{Performance of DBNN on set $\mathcal{B}$}\label{tab:results_B}
		\begin{tabular}{llllllll}
			\toprule
			$n$       &   $A$ &     $TPR$ &     $PPV$ &     $TNR$ &     $NPV$ &      $F_1$ &   $AUROC$         \\
			\midrule
			$5$       & $0.781$ & $0.790$ & $0.776$ & $0.772$ & $0.786$ & $0.783$ & $0.868$ \\
			$15$      & $0.871$ & $0.792$ & $0.942$ & $0.951$ & $0.820$ & $0.860$ & $0.941$ \\
			$25$      & $0.885$ & $0.843$ & $0.921$ & $0.928$ & $0.855$ & $0.880$ & $0.961$ \\
			$35$      & $0.911$ & $0.890$ & $0.928$ & $0.931$ & $0.895$ & $0.909$ & $0.976$ \\
			$45$      & $0.905$ & $0.879$ & $0.926$ & $0.930$ & $0.885$ & $0.902$ & $0.974$ \\
			$55$      & $0.907$ & $0.872$ & $0.937$ & $0.941$ & $0.880$ & $0.903$ & $0.979$ \\
			$65$      & $0.926$ & $0.900$ & $0.949$ & $0.952$ & $0.905$ & $0.924$ & $0.986$ \\
			$75$      & $0.925$ & $0.886$ & $0.961$ & $0.964$ & $0.894$ & $0.922$ & $0.986$ \\
			$85$      & $0.927$ & $0.907$ & $0.944$ & $0.946$ & $0.911$ & $0.925$ & $0.986$ \\
			$95$      & $0.933$ & $0.902$ & $0.962$ & $0.964$ & $0.908$ & $0.931$ & $0.988$ \\
			\midrule
			overall & $0.897$ & $0.866$ & $0.923$ & $0.928$ & $0.874$ & $0.894$ & $0.969$ \\
			\bottomrule
		\end{tabular}
	\end{table}
	
	\subsection{Results of Comparison with Other Normality Tests on Sets $\mathcal{C}$ and $\mathcal{D}$}\label{subsec:comparison}
		
\begin{figure}
		\subfloat[$G_1$]{\includegraphics[width = 2.5in]{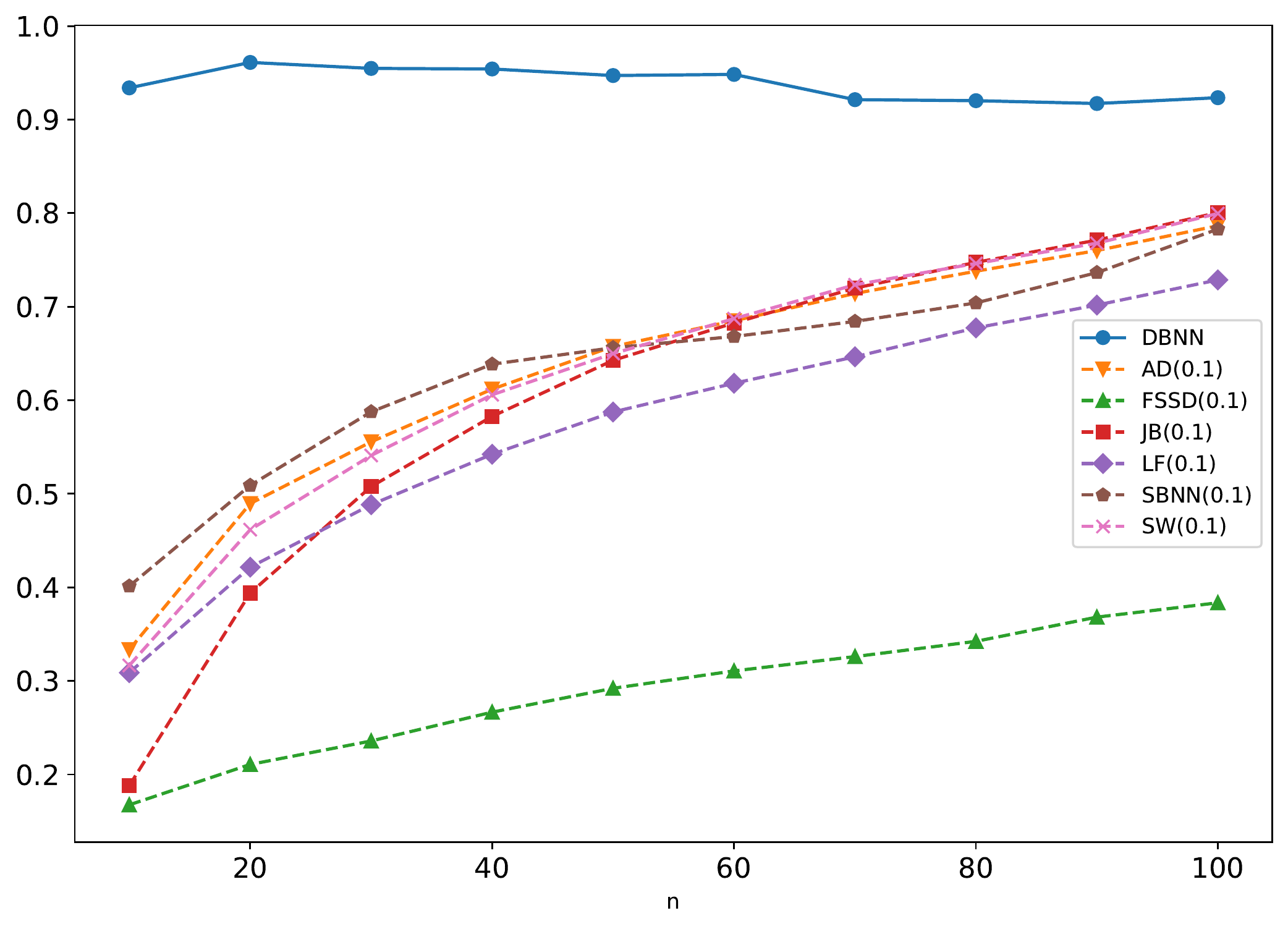}} 
		\subfloat[$G_2$]{\includegraphics[width = 2.5in]{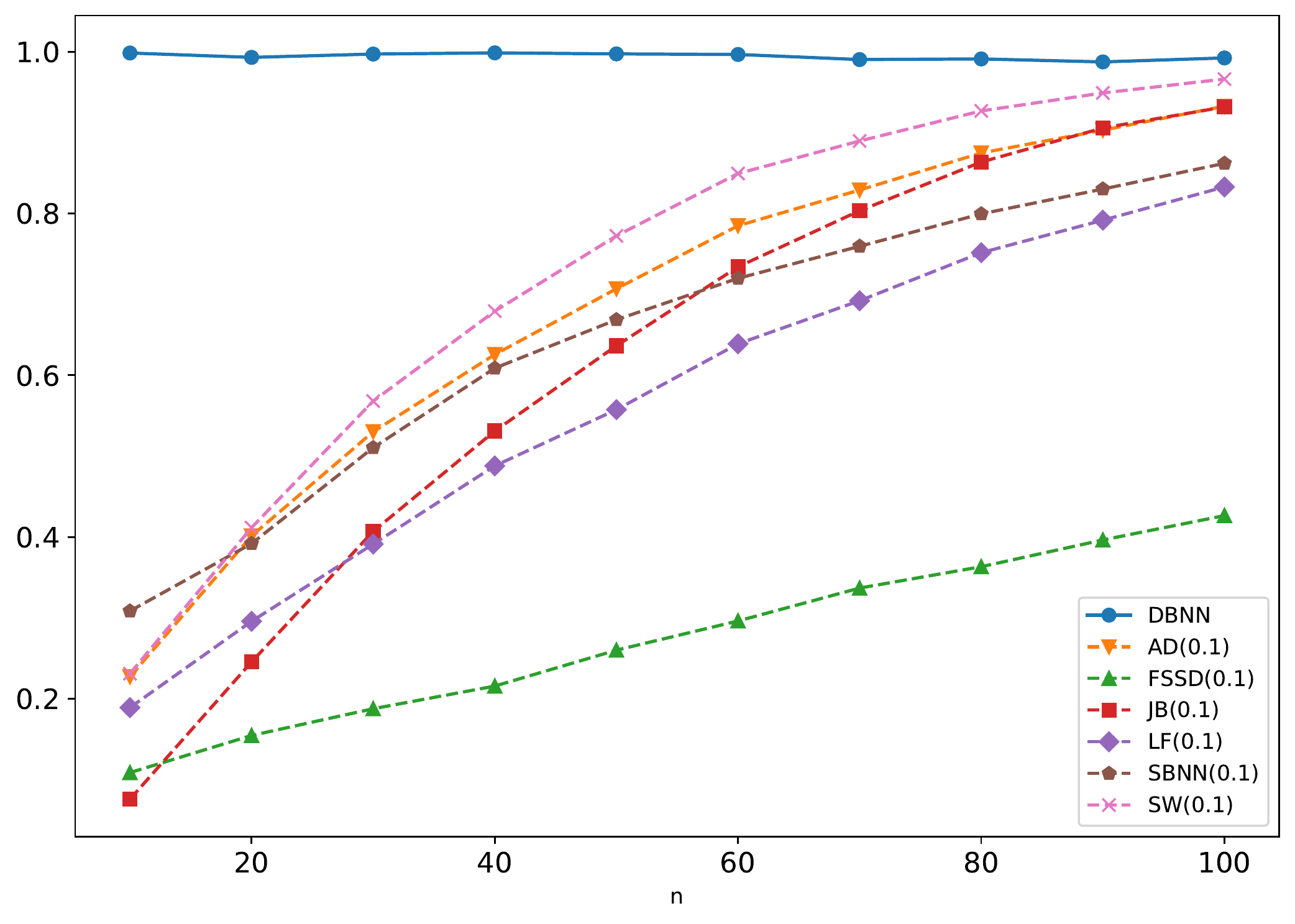}} \\
		\subfloat[$G_3$]{\includegraphics[width = 2.5in]{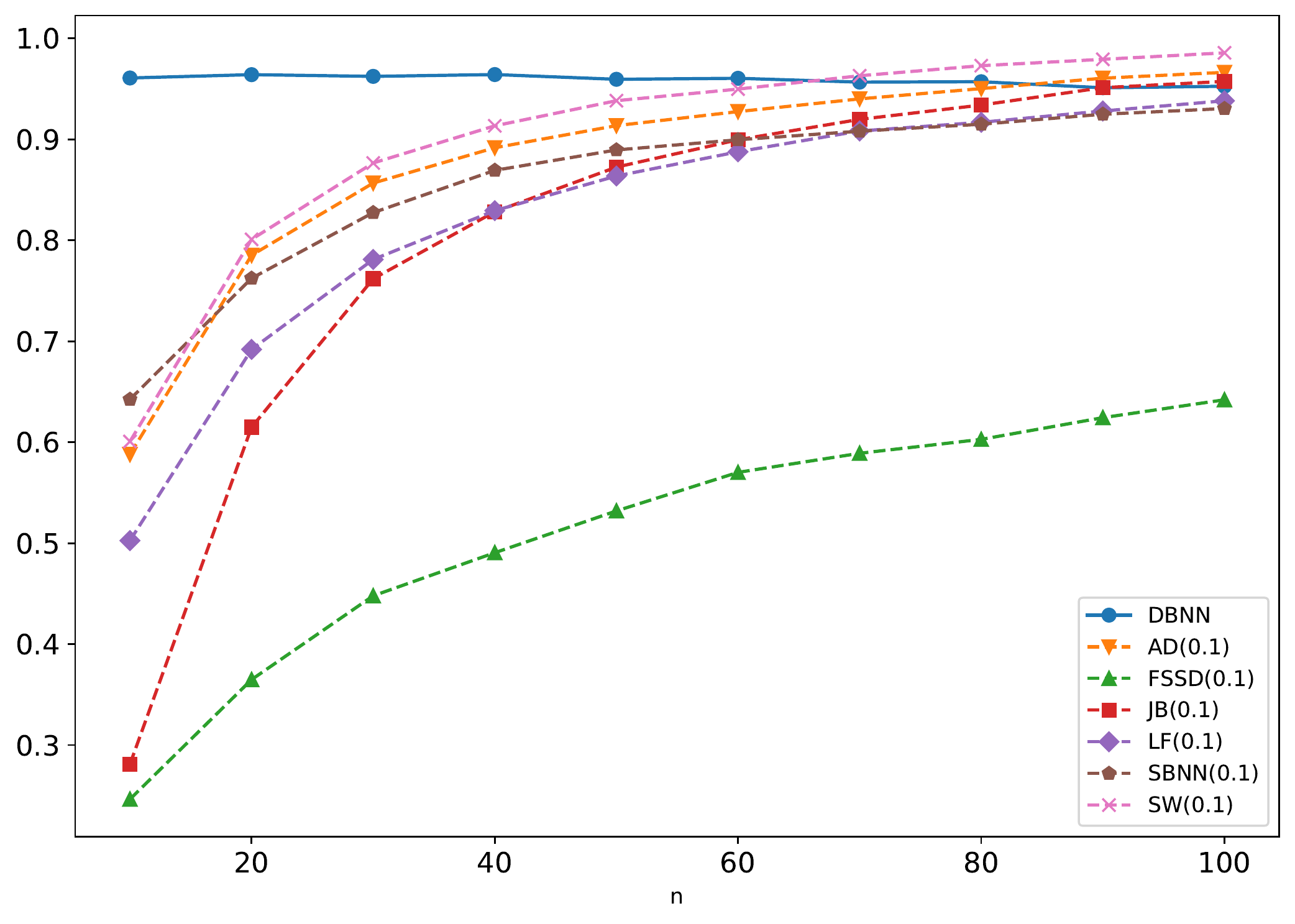}} 
		\subfloat[$G_4$]{\includegraphics[width = 2.5in]{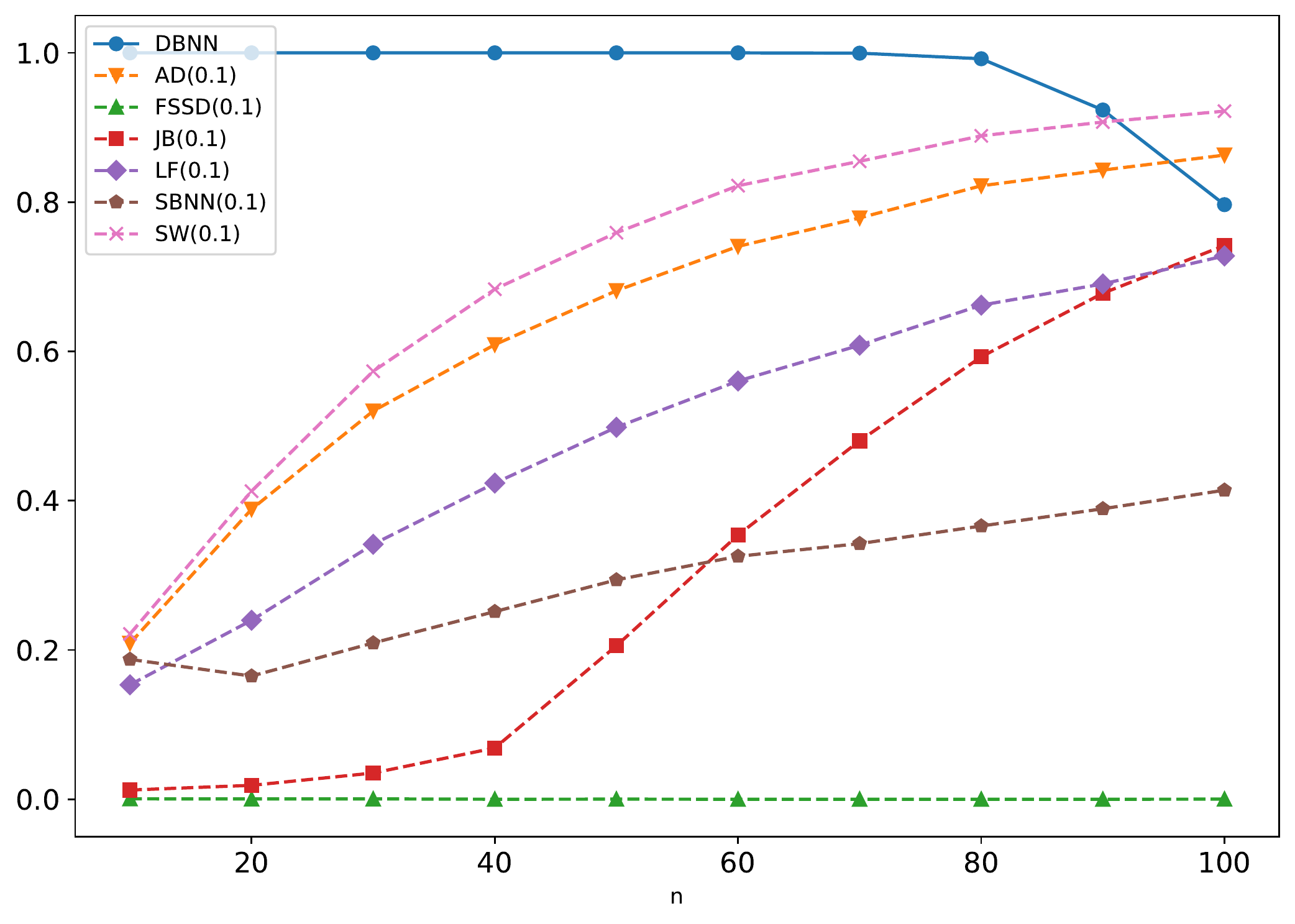}}  
		\caption{$TNR$ of DBNN, SBNN and tests AD, JB, LF, SW, and FSSD at the nominal level $\alpha=0.1$ on the samples from $G_1,G_2,G_3,G_4$. The DBNN's $TNR$ is consistently high, in almost all the cases better than those of other approaches, with the exception of the samples with $100$ elements from $G_4$, for which DBNN has the third best $TNR$ score.}
		\label{fig:power_per_group}
	\end{figure}

	We used set $\mathcal{C}$ to evaluate statistical power ($TNR$) of DBNN and compare it to the selected statistical tests of normality and SBNN of \cite{Sigut2006}. The nominal Type I error rate for all the tests was set to $\alpha=0.1$. For each statistical test, it holds that for any $\alpha \in (0, 1)$, the test will have $TPR \approx 1 - \alpha$ when evaluated on a dataset that is large enough. The tests with higher $\alpha$ are more powerful, the price of which is a greater Type I error rate. The nominal level of $0.1$ is greater than the most common nominal levels of $0.01$ and $0.05$, so the powers of the tests at that level, estimated in this simulation, exceed the powers that those tests have in practice. The results for all the groups $G_1-G_4$ from set $\mathcal{C}$ are presented in Tables \ref{tab:G1}-\ref{tab:G4} and visualized in Figure \ref{fig:power_per_group}.  Since $\mathcal{C}$ contains only non-normal samples, $NPV$ is trivially equal to $1$ for all the methods tested. In terms of $TNR$, DBNN clearly outperformed all the other methods.
	
	\begin{table}
		\centering
		\caption{$TNR$ of DBNN, SBNN and tests AD, JB, LF, SW, and FSSD at the nominal level $\alpha=0.1$ on the samples from $G_1$.}
		\label{tab:G1}
		\begin{tabular}{llllllll}
			\toprule
			$n$ &    DBNN &  AD &  FSSD &  JB &  LF &  SBNN &  SW \\
			\midrule
			$10$      & $0.934$ &  $0.333$ &    $0.167$ &  $0.188$ &  $0.309$ &    $0.401$ &  $0.316$ \\
			$20$      & $0.961$ &  $0.490$ &    $0.211$ &  $0.394$ &  $0.422$ &    $0.509$ &  $0.462$ \\
			$30$      & $0.955$ &  $0.556$ &    $0.236$ &  $0.508$ &  $0.488$ &    $0.588$ &  $0.541$ \\
			$40$      & $0.954$ &  $0.612$ &    $0.267$ &  $0.583$ &  $0.542$ &    $0.639$ &  $0.606$ \\
			$50$      & $0.947$ &  $0.658$ &    $0.292$ &  $0.643$ &  $0.587$ &    $0.656$ &  $0.650$ \\
			$60$      & $0.948$ &  $0.685$ &    $0.311$ &  $0.683$ &  $0.618$ &    $0.668$ &  $0.687$ \\
			$70$      & $0.921$ &  $0.714$ &    $0.326$ &  $0.720$ &  $0.646$ &    $0.684$ &  $0.723$ \\
			$80$      & $0.920$ &  $0.738$ &    $0.342$ &  $0.747$ &  $0.677$ &    $0.704$ &  $0.746$ \\
			$90$      & $0.917$ &  $0.760$ &    $0.368$ &  $0.771$ &  $0.702$ &    $0.736$ &  $0.768$ \\
			$100$     & $0.923$ &  $0.787$ &    $0.383$ &  $0.801$ &  $0.729$ &    $0.783$ &  $0.800$ \\
			\midrule
			overall & $0.938$ &  $0.633$ &    $0.290$ &  $0.604$ &  $0.572$ &    $0.637$ &  $0.630$ \\
			\bottomrule
		\end{tabular}
	\end{table}
	\begin{table}
		\centering
		\caption{$TNR$ of DBNN, SBNN and tests AD, JB, LF, SW, and FSSD at the nominal level $\alpha=0.1$ on the samples from $G_2$.}
		\label{tab:G2}
		\begin{tabular}{llllllll}
			\toprule
			$n$ &    DBNN &  AD &  FSSD &  JB &  LF &  SBNN &  SW \\
			\midrule
			$10$      & $0.998$ &  $0.227$ &    $0.109$ &  $0.076$ &  $0.189$ &    $0.308$ &  $0.231$ \\
			$20$      & $0.993$ &  $0.401$ &    $0.155$ &  $0.246$ &  $0.296$ &    $0.392$ &  $0.412$ \\
			$30$      & $0.997$ &  $0.530$ &    $0.188$ &  $0.407$ &  $0.391$ &    $0.510$ &  $0.568$ \\
			$40$      & $0.999$ &  $0.626$ &    $0.216$ &  $0.531$ &  $0.488$ &    $0.609$ &  $0.679$ \\
			$50$      & $0.997$ &  $0.707$ &    $0.260$ &  $0.636$ &  $0.557$ &    $0.669$ &  $0.772$ \\
			$60$      & $0.997$ &  $0.785$ &    $0.296$ &  $0.734$ &  $0.639$ &    $0.720$ &  $0.849$ \\
			$70$      & $0.990$ &  $0.829$ &    $0.337$ &  $0.803$ &  $0.692$ &    $0.759$ &  $0.890$ \\
			$80$      & $0.991$ &  $0.875$ &    $0.363$ &  $0.864$ &  $0.752$ &    $0.800$ &  $0.927$ \\
			$90$      & $0.987$ &  $0.903$ &    $0.396$ &  $0.906$ &  $0.792$ &    $0.830$ &  $0.949$ \\
			$100$     & $0.992$ &  $0.933$ &    $0.426$ &  $0.932$ &  $0.833$ &    $0.862$ &  $0.966$ \\
			overall & $0.994$ &  $0.682$ &    $0.275$ &  $0.613$ &  $0.563$ &    $0.646$ &  $0.724$ \\
			\bottomrule
		\end{tabular}
	\end{table}
	
	\begin{table}
		\centering
		\caption{$TNR$ of DBNN, SBNN and tests AD, JB, LF, SW, and FSSD at the nominal level $\alpha=0.1$ on the samples from $G_3$.}
		\label{tab:G3}
		\begin{tabular}{llllllll}
			\toprule
			$n$ &    DBNN &  AD &  FSSD &  JB &  LF &  SBNN &  SW \\
			\midrule
			$10$      & $0.961$ &  $0.588$ &    $0.246$ &  $0.281$ &  $0.503$ &    $0.642$ &  $0.601$ \\
			$20$      & $0.964$ &  $0.785$ &    $0.365$ &  $0.615$ &  $0.692$ &    $0.762$ &  $0.801$ \\
			$30$      & $0.962$ &  $0.857$ &    $0.448$ &  $0.762$ &  $0.781$ &    $0.827$ &  $0.877$ \\
			$40$      & $0.964$ &  $0.892$ &    $0.491$ &  $0.828$ &  $0.829$ &    $0.869$ &  $0.913$ \\
			$50$      & $0.960$ &  $0.914$ &    $0.532$ &  $0.873$ &  $0.864$ &    $0.890$ &  $0.938$ \\
			$60$      & $0.961$ &  $0.928$ &    $0.570$ &  $0.900$ &  $0.888$ &    $0.900$ &  $0.950$ \\
			$70$      & $0.957$ &  $0.940$ &    $0.589$ &  $0.920$ &  $0.908$ &    $0.908$ &  $0.963$ \\
			$80$      & $0.957$ &  $0.950$ &    $0.603$ &  $0.934$ &  $0.917$ &    $0.915$ &  $0.973$ \\
			$90$      & $0.951$ &  $0.961$ &    $0.624$ &  $0.951$ &  $0.928$ &    $0.925$ &  $0.979$ \\
			$100$     & $0.953$ &  $0.966$ &    $0.642$ &  $0.957$ &  $0.938$ &    $0.931$ &  $0.986$ \\
			\midrule
			overall & $0.959$ &  $0.878$ &    $0.511$ &  $0.802$ &  $0.825$ &    $0.857$ &  $0.898$ \\
			\bottomrule
		\end{tabular}
	\end{table}
	
	\begin{table}
		\centering
		\caption{$TNR$ of DBNN, SBNN and tests AD, JB, LF, SW, and FSSD at the nominal level $\alpha=0.1$ on the samples from $G_4$.}
		\label{tab:G4}
		\begin{tabular}{llllllll}
			\toprule
			$n$ &    DBNN &  AD &  FSSD &  JB &  LF &  SBNN &  SW \\
			\midrule
			$10$      & $1.000$ &  $0.209$ &    $0.000$ &  $0.012$ &  $0.153$ &    $0.187$ &  $0.221$ \\
			$20$      & $1.000$ &  $0.389$ &    $0.000$ &  $0.019$ &  $0.240$ &    $0.165$ &  $0.413$ \\
			$30$      & $1.000$ &  $0.520$ &    $0.000$ &  $0.035$ &  $0.342$ &    $0.209$ &  $0.573$ \\
			$40$      & $1.000$ &  $0.609$ &    $0.000$ &  $0.069$ &  $0.424$ &    $0.251$ &  $0.683$ \\
			$50$      & $1.000$ &  $0.681$ &    $0.000$ &  $0.206$ &  $0.498$ &    $0.294$ &  $0.759$ \\
			$60$      & $1.000$ &  $0.741$ &    $0.000$ &  $0.354$ &  $0.560$ &    $0.325$ &  $0.822$ \\
			$70$      & $1.000$ &  $0.779$ &    $0.000$ &  $0.480$ &  $0.608$ &    $0.342$ &  $0.855$ \\
			$80$      & $0.992$ &  $0.822$ &    $0.000$ &  $0.593$ &  $0.662$ &    $0.366$ &  $0.889$ \\
			$90$      & $0.923$ &  $0.843$ &    $0.000$ &  $0.678$ &  $0.690$ &    $0.389$ &  $0.907$ \\
			$100$     & $0.797$ &  $0.863$ &    $0.000$ &  $0.742$ &  $0.728$ &    $0.414$ &  $0.922$ \\
			\midrule
			overall & $0.971$ &  $0.646$ &    $0.000$ &  $0.319$ &  $0.491$ &    $0.294$ &  $0.704$ \\
			\bottomrule
		\end{tabular}
	\end{table}
	
\begin{figure}
		\centering
		\caption{The ROC-curves of DBNN, SBNN, and the tests SW, LF, JB, AD, and FSSD, determined using set $\mathcal{D}$}\label{fig:roc_curves}
		\includegraphics[width=0.7\textwidth]{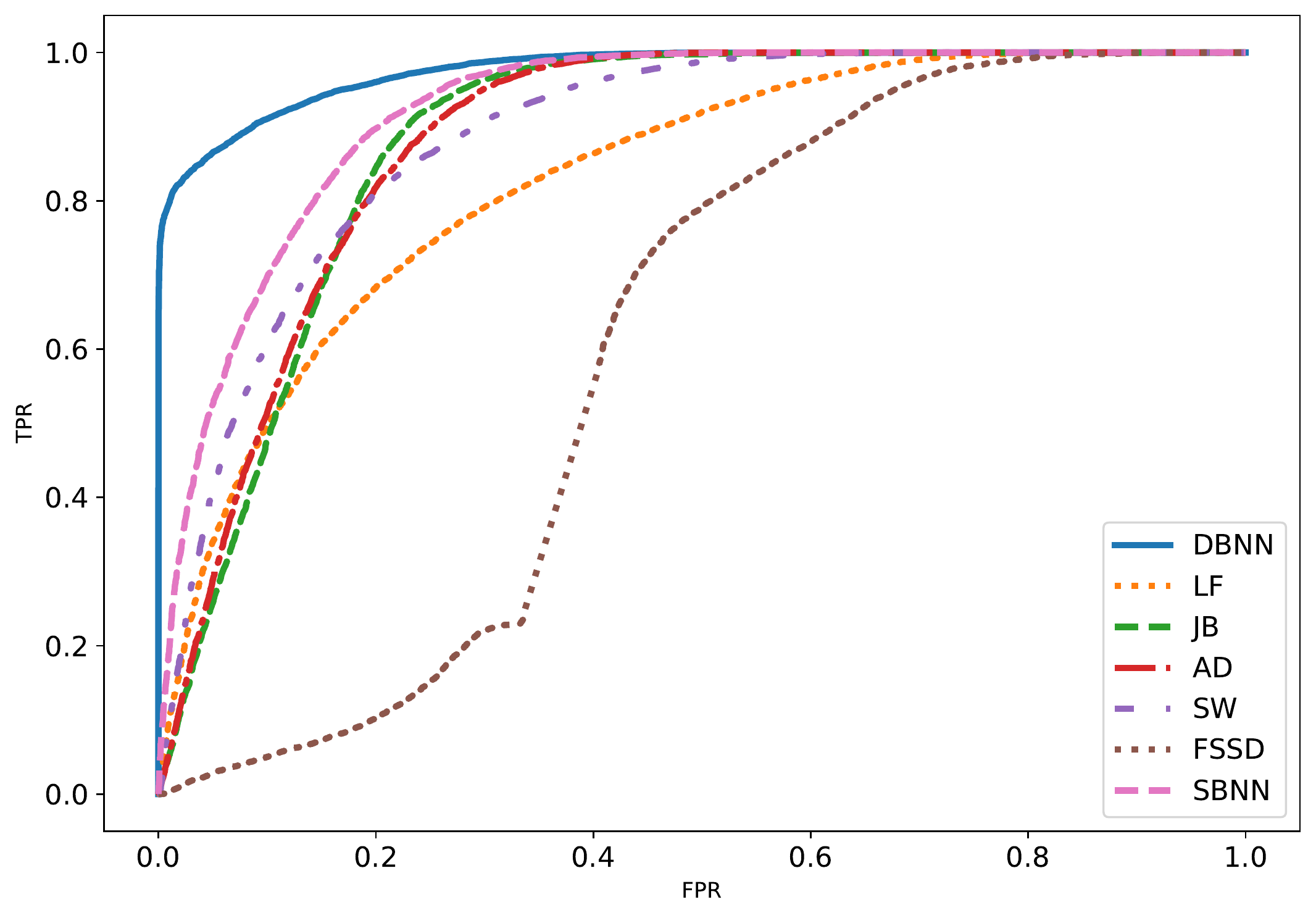}
	\end{figure}

	We calculated $A$, $TPR$, $PPV$, $TNR$, $NPV$, and $F_1$ for DBNN, SBNN and tests SW, LF, JB, AD and FSSD with nominal $\alpha=0.01, 0.05, 0.1$ on set $\mathcal{D}$. The results are presented in Table \ref{tab:results_D}. The values of $AUROC$ are in Table \ref{tab:auroc}, and the ROC curves are visualized in Figure \ref{fig:roc_curves}. DBNN's $AUROC$ is exceptionally high and almost equal to $1$, the score that corresponds to a perfect classifier.
	
	\begin{table}
		\centering
		\caption{Performance of DBNN, SBNN and tests AD, JB, LF, SW, FSSD with nominal $\alpha=0.01, 0.05, 0.1$ on the samples from $\mathcal{D}$.}
		\label{tab:results_D}
		\begin{tabular}{ll|llllll}
			\toprule
			&  $\alpha$   &       $A$ &    $TPR$ &     $PPV$ &     $TNR$ &     $NPV$ &      $F_1$ \\
			\midrule
			SW & $0.01$ & $0.819$ & $0.990$ & $0.738$ & $0.649$ & $0.985$ & $0.846$  \\
			& $0.05$ & $0.839$ & $0.952$ & $0.776$ & $0.726$ & $0.938$ & $0.855$  \\
			& $0.1$ & $0.834$ & $0.903$ & $0.794$ & $0.766$ & $0.887$ & $0.845$  \\
			LF & $0.01$ & $0.759$ & $0.991$ & $0.677$ & $0.527$ & $0.984$ & $0.804$ \\
			& $0.05$ & $0.792$ & $0.952$ & $0.721$ & $0.631$ & $0.929$ & $0.820$  \\
			& $0.1$ & $0.797$ & $0.902$ & $0.746$ & $0.692$ & $0.876$ & $0.816$ \\
			JB & $0.01$ & $0.810$ & $0.986$ & $0.730$ & $0.635$ & $0.978$ & $0.839$  \\
			& $0.05$ & $0.829$ & $0.968$ & $0.758$ & $0.690$ & $0.955$ & $0.850$  \\
			& $0.1$ & $0.836$ & $0.949$ & $0.774$ & $0.723$ & $0.934$ & $0.853$  \\
			AD & $0.01$ & $0.800$ & $0.990$ & $0.718$ & $0.611$ & $0.984$ & $0.832$  \\
			& $0.05$ & $0.825$ & $0.949$ & $0.761$ & $0.702$ & $0.932$ & $0.845$  \\
			& $0.1$ & $0.825$ & $0.897$ & $0.784$ & $0.753$ & $0.879$ & $0.837$  \\
			FSSD & $0.01$ & $0.487$ & $0.691$ & $0.494$ & $0.279$ & $0.470$ & $0.576$  \\
			& $0.05$ & $0.504$ & $0.676$ & $0.506$ & $0.329$ & $0.499$ & $0.579$ \\
			& $0.1$ & $0.517$ & $0.648$ & $0.517$ & $0.383$ & $0.517$ & $0.575$  \\
			SBNN & /    & $0.847$ & $0.931$ & $0.797$ & $0.763$ & $0.918$ & $0.859$  \\
			DBNN & /    & $0.907$ & $0.868$ & $0.941$ & $0.946$ & $0.878$ & $0.903$  \\
			\bottomrule
		\end{tabular}
	\end{table}
	
	\begin{table}
		\centering
		\caption{The $AUROC$ of DBNN, SBNN, and the tests SW, LF, JB, AD, and FSSD, calculated using set $\mathcal{D}$}\label{tab:auroc}
		\begin{tabular}{llllllll}
			\toprule
			& DBNN & SBNN & SW & LF & JB & AD & FSSD \\
			\midrule
			$AUROC$ & $0.978$ & $0.921$ & $0.886$ & $0.829$ & $0.882$ & $0.883$ & $0.604$ \\
			\bottomrule
		\end{tabular}
	\end{table}
	
	\subsection{Evaluation on Real-World Data}\label{subsec:evaluation_on_real_world_data}
	The results on the height data from set $\mathcal{R}_{height}$ are presented in Table \ref{tab:height_results}. DBNN's TPR is comparable to those of standard tests at $\alpha=0.01$ for samples with more than $n=10$ elements. For samples with fewer than $10$ elements, DBNN performs much worse.
	
	\begin{table}
		\centering
		\caption{TPR of DBNN, SBNN, SW, LG, JB, AD, and FSSD on the height dataset $\mathcal{R}_{height}$. DBNN is very accurate for the samples with $n\geq10$ elements, but not so much for those with fewer than $10$ elements.}
		\label{tab:height_results}
		\begin{tabular}{llll}
			\toprule
			&   $\alpha$  &  $n < 10$ &  $n \geq 10$ \\
			\midrule
			DBNN & / &   $0.571$ &      $0.979$ \\
			SW & 0.01 &   $1.000$ &      $0.979$ \\
			& 0.05 &   $1.000$ &      $0.906$ \\
			& 0.1 &   $1.000$ &      $0.854$ \\
			LF & 0.01 &   $0.714$ &      $0.990$ \\
			& 0.05 &   $0.643$ &      $0.906$ \\
			& 0.1 &   $0.643$ &      $0.812$ \\
			JB & 0.01 &   $1.000$ &      $0.906$ \\
			& 0.05 &   $1.000$ &      $0.896$ \\
			& 0.1 &   $1.000$ &      $0.896$ \\
			AD & 0.01 &   $0.893$ &      $0.990$ \\
			& 0.05 &   $0.893$ &      $0.938$ \\
			& 0.1 &   $0.893$ &      $0.833$ \\
			FSSD & 0.01 &   $0.964$ &      $0.958$ \\
			& 0.05 &   $1.000$ &      $0.938$ \\
			& 0.1 &   $0.893$ &      $0.896$ \\
			SBNN &  /   &   $0.750$ &      $0.792$ \\
			\bottomrule
		\end{tabular}
	\end{table}

	The results on the non-normal samples from set $\mathcal{R}_{earthquake}$ are presented in Table \ref{tab:earthquake_results}. Except in the case of $7$ out of $1000$ samples with $5$ elements, DBNN successfully identified non-normality in all the samples, achieving TNR of $100\%$ for samples with $10$ or more elements, and $99.3\%$ for the samples with $5$ elements.
	
	\begin{landscape}
	\begin{table}
		\centering
		\caption{TNR of DBNN, SW, LG, JB, AD, FSSD, and SBNN on set $\mathcal{R}_{earthquake}$. DBNN clearly beats SBNN and the standard tests, especially in the case of very small samples. }
		\label{tab:earthquake_results}
		\begin{tabular}{llllllllllllllllll}
			\toprule
			 &    DBNN & \multicolumn{3}{l}{SW} & \multicolumn{3}{l}{LF} & \multicolumn{3}{l}{JB} & \multicolumn{3}{l}{AD} & \multicolumn{3}{l}{FSSD} &    SBNN \\
			$n$ &{/} &    0.01 &    0.05 &     0.1 &    0.01 &    0.05 &     0.1 &    0.01 &    0.05 &     0.1 &    0.01 &    0.05 &     0.1 &    0.01 &    0.05 & \multicolumn{2}{l}{0.1} \\
			\midrule
			$5$   & $0.993$ & $0.057$ & $0.155$ & $0.246$ & $0.049$ & $0.131$ & $0.233$ & $0.000$ & $0.000$ & $0.000$ & $0.000$ & $0.013$ & $0.051$ & $0.019$ & $0.036$ & $0.049$ & $0.538$ \\
			$10$  & $1.000$ & $0.231$ & $0.443$ & $0.569$ & $0.117$ & $0.315$ & $0.437$ & $0.077$ & $0.138$ & $0.185$ & $0.196$ & $0.415$ & $0.552$ & $0.038$ & $0.055$ & $0.078$ & $0.626$ \\
			$15$  & $1.000$ & $0.421$ & $0.680$ & $0.791$ & $0.227$ & $0.467$ & $0.589$ & $0.212$ & $0.312$ & $0.381$ & $0.422$ & $0.652$ & $0.764$ & $0.055$ & $0.090$ & $0.137$ & $0.745$ \\
			$20$  & $1.000$ & $0.615$ & $0.840$ & $0.911$ & $0.310$ & $0.573$ & $0.696$ & $0.345$ & $0.456$ & $0.545$ & $0.585$ & $0.804$ & $0.878$ & $0.045$ & $0.101$ & $0.158$ & $0.827$ \\
			$25$  & $1.000$ & $0.761$ & $0.918$ & $0.966$ & $0.430$ & $0.687$ & $0.795$ & $0.457$ & $0.587$ & $0.679$ & $0.706$ & $0.881$ & $0.939$ & $0.076$ & $0.152$ & $0.229$ & $0.891$ \\
			$30$  & $1.000$ & $0.857$ & $0.964$ & $0.986$ & $0.508$ & $0.764$ & $0.867$ & $0.504$ & $0.649$ & $0.764$ & $0.808$ & $0.933$ & $0.968$ & $0.082$ & $0.163$ & $0.261$ & $0.930$ \\
			$35$  & $1.000$ & $0.939$ & $0.987$ & $0.990$ & $0.668$ & $0.860$ & $0.934$ & $0.657$ & $0.807$ & $0.885$ & $0.895$ & $0.973$ & $0.983$ & $0.104$ & $0.211$ & $0.360$ & $0.966$ \\
			$40$  & $1.000$ & $0.965$ & $1.000$ & $1.000$ & $0.698$ & $0.903$ & $0.957$ & $0.697$ & $0.825$ & $0.908$ & $0.938$ & $0.985$ & $0.994$ & $0.112$ & $0.237$ & $0.366$ & $0.979$ \\
			$45$  & $1.000$ & $0.993$ & $0.999$ & $1.000$ & $0.805$ & $0.941$ & $0.976$ & $0.790$ & $0.909$ & $0.963$ & $0.971$ & $0.996$ & $0.998$ & $0.143$ & $0.312$ & $0.471$ & $0.995$ \\
			$50$  & $1.000$ & $0.995$ & $1.000$ & $1.000$ & $0.863$ & $0.975$ & $0.991$ & $0.848$ & $0.943$ & $0.976$ & $0.985$ & $0.997$ & $1.000$ & $0.186$ & $0.361$ & $0.521$ & $0.995$ \\
			$55$  & $1.000$ & $0.996$ & $0.999$ & $1.000$ & $0.883$ & $0.975$ & $0.990$ & $0.871$ & $0.961$ & $0.987$ & $0.988$ & $0.998$ & $0.998$ & $0.175$ & $0.358$ & $0.557$ & $0.998$ \\
			$60$  & $1.000$ & $0.999$ & $1.000$ & $1.000$ & $0.936$ & $0.987$ & $0.997$ & $0.921$ & $0.985$ & $0.995$ & $0.994$ & $1.000$ & $1.000$ & $0.209$ & $0.418$ & $0.585$ & $0.997$ \\
			$65$  & $1.000$ & $1.000$ & $1.000$ & $1.000$ & $0.971$ & $0.998$ & $0.999$ & $0.945$ & $0.995$ & $0.999$ & $1.000$ & $1.000$ & $1.000$ & $0.263$ & $0.487$ & $0.661$ & $1.000$ \\
			$70$  & $1.000$ & $1.000$ & $1.000$ & $1.000$ & $0.978$ & $0.998$ & $0.998$ & $0.962$ & $0.996$ & $0.999$ & $0.998$ & $1.000$ & $1.000$ & $0.286$ & $0.550$ & $0.700$ & $1.000$ \\
			$75$  & $1.000$ & $1.000$ & $1.000$ & $1.000$ & $0.987$ & $0.997$ & $0.999$ & $0.981$ & $0.997$ & $0.999$ & $0.998$ & $0.999$ & $1.000$ & $0.322$ & $0.580$ & $0.748$ & $1.000$ \\
			$80$  & $1.000$ & $1.000$ & $1.000$ & $1.000$ & $0.987$ & $0.999$ & $1.000$ & $0.985$ & $1.000$ & $1.000$ & $1.000$ & $1.000$ & $1.000$ & $0.352$ & $0.608$ & $0.787$ & $1.000$ \\
			$85$  & $1.000$ & $1.000$ & $1.000$ & $1.000$ & $0.996$ & $1.000$ & $1.000$ & $0.993$ & $0.999$ & $1.000$ & $1.000$ & $1.000$ & $1.000$ & $0.371$ & $0.654$ & $0.790$ & $1.000$ \\
			$90$  & $1.000$ & $1.000$ & $1.000$ & $1.000$ & $0.997$ & $1.000$ & $1.000$ & $0.999$ & $1.000$ & $1.000$ & $1.000$ & $1.000$ & $1.000$ & $0.455$ & $0.714$ & $0.846$ & $1.000$ \\
			$95$  & $1.000$ & $1.000$ & $1.000$ & $1.000$ & $0.994$ & $0.998$ & $0.999$ & $0.997$ & $0.999$ & $0.999$ & $0.999$ & $1.000$ & $1.000$ & $0.455$ & $0.710$ & $0.860$ & $1.000$ \\
			$100$ & $1.000$ & $1.000$ & $1.000$ & $1.000$ & $0.999$ & $1.000$ & $1.000$ & $0.999$ & $1.000$ & $1.000$ & $1.000$ & $1.000$ & $1.000$ & $0.495$ & $0.751$ & $0.889$ & $1.000$ \\
			\bottomrule
		\end{tabular}
	\end{table}
	\end{landscape}
	
	\subsection{Runtime Analysis}\label{subsec:runtime_analysis}
	
	Using random sets consisting of $10\%, 32.5\%, 55\% , 77.5\%$, and $100\%$ of $\mathcal{A}_{cv}$, we trained and tested a network with the same design as our DBNN ($100$ and $10$ neurons in two hidden layers, $q=0.1$, $c=0.1$) twenty times, partitioning the subset in question into twenty folds, training the network on nineteen folds, and testing it on the remaining one.
	
\begin{figure}
		\centering
		\caption{The median learning curves of DBNN on $\mathcal{A}_{cv}$, determined via $20$-fold cross-validation}\label{fig:learning_curves}
		\includegraphics[width=0.7\textwidth]{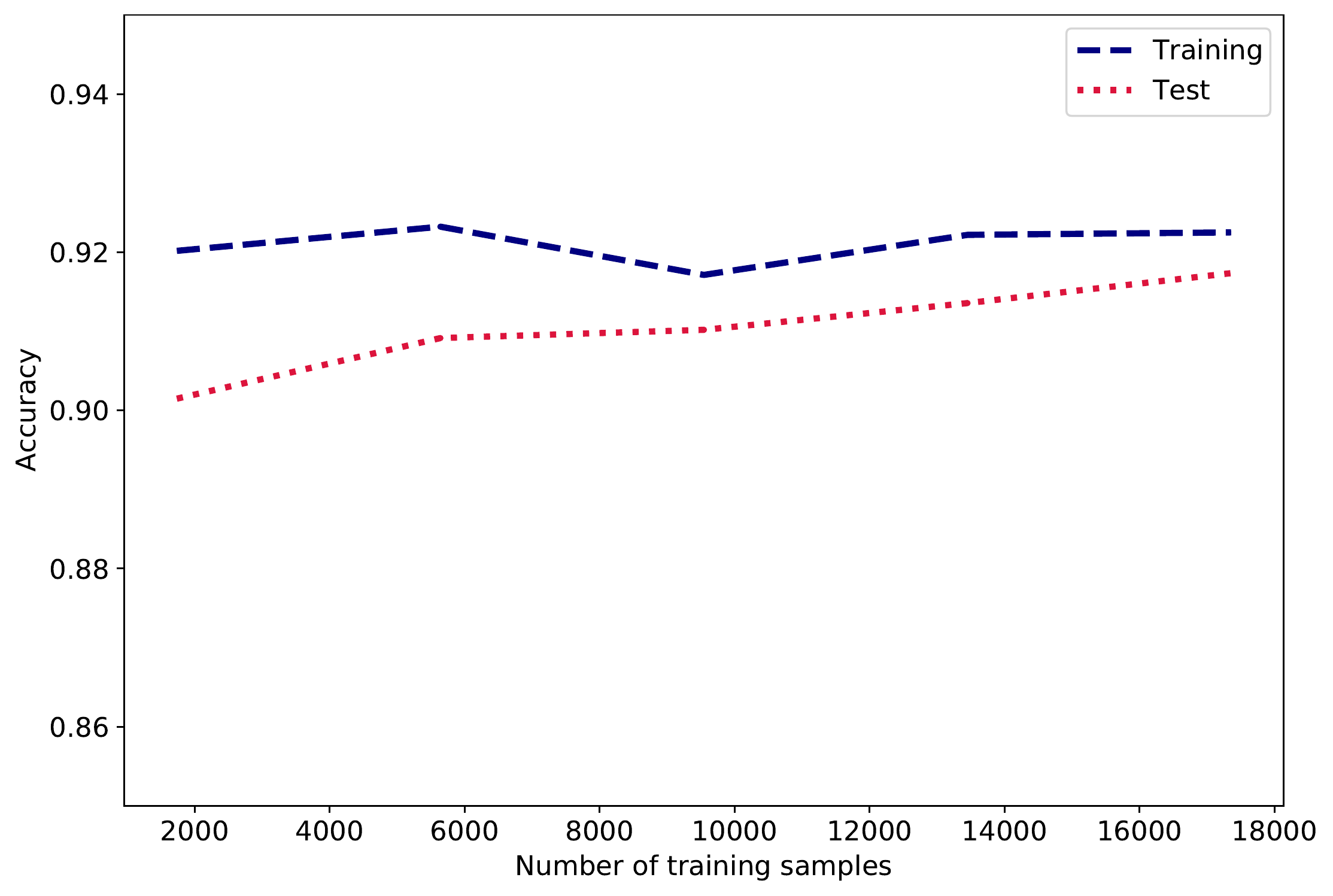}
	\end{figure}

	Figure \ref{fig:learning_curves} presents the medians of the training and test accuracy scores calculated that way. We see that the test score increases with the size of the training set and that the median training and test accuracies are very similar. This indicates that the network does not overfit the data. The median test accuracy is around $91\%$, which is a good score, so the network does not underfit the data either.
	
\begin{figure}
		\centering
		\caption{Fit times of DBNN on the increasing-size subsets of $\mathcal{A}_{cv}$. The solid line connects the median times for each training set size. The circles represent the times of individual fits.}\label{fig:fit_times}
		\includegraphics[width=0.7\textwidth]{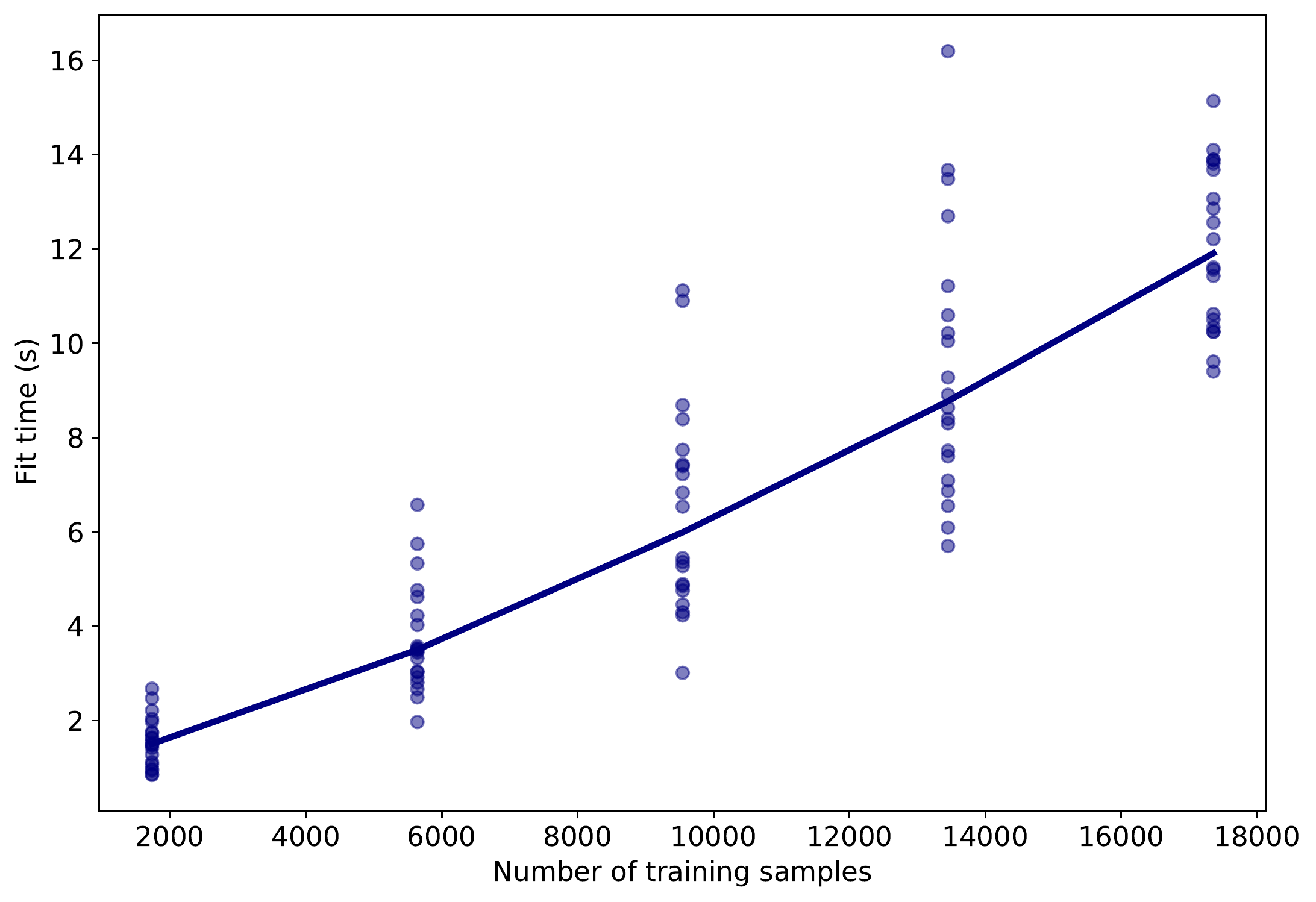}
	\end{figure}

	The relationship between training time and the size of the training dataset is depicted in Figure \ref{fig:fit_times}. As expected, the time needed to train the network increases as more and more data are used for training. The variance of the fit times appears to be higher for the sets with more than $30\%$ of $\mathcal{A}_{cv}$ than for the set which contains only $10\%$ of $\mathcal{A}_{cv}$. Still, all the times are between $1$ and $16$ seconds. The network can be trained very quickly. 

\section{Discussion}\label{sec:discussion}

In this study, we created an artificial neural networks capable of testing for normality of a distribution using only a small sample drawn from it. The network, named the Descriptor-Based Neural Network (DBNN), proved competitive with standard statistical tests of normality on the samples with $100$ elements at most, exceeding them in many performance metrics, especially statistical power.

Numerical experiments conducted on the samples from various normal and non-normal distributions indicated that DBNN was generally more accurate than standard normality tests. In particular, DBNN appears to have higher statistical power (true negative rate, $TNR$), i.e. it is better at detecting non-normal distributions than those tests. The power of DBNN was measured to be almost $100\%$ even for the samples with only $10$ elements, which is the level of power that standard tests do not reach even on the samples with more than $100$ elements at a very high nominal Type I error rate of $\alpha=10\%$. However, on artificial datasets, DBNN had a lower true positive rate ($TPR$, the complement of Type I error rate) than those tests. That means that DBNN is worse at detecting normal distributions. This was especially true for very small samples in sets $\mathcal{A}_{test}, \mathcal{B}$, and $\mathcal{D}$. However, in the same sets, as the sample size grew, so did the DBNN's $TPR$. DBNN had a higher positive predictive power: there were more normal samples among those that DBNN  classified as such than among those classified as normal by the standard tests. Conversely, it showed a lower negative predictive value. Still, the trade-off between DBNN's $TPR$ and $TNR$ appears to be such that, when the overall measures such as classification accuracy, the $F_1$ score and $AUROC$ are taken into account, the DBNN seems to perform better than the tests with which we compared it. Its $AUROC$ was estimated to be $0.978$, which is almost equal to the theoretical maximal $AUROC$ of $1$ that corresponds to perfect classifiers. This indicates that DBNN does not simply trade off $TPR$ for $TNR$: the network is characterized by a better classification accuracy than the tests considered, which is a conclusion supported by the ROC curves in Figure \ref{fig:roc_curves}. A possible explanation is as follows. The neural network learns efficient classification rules directly from the data, whereas the tests are constructed with a specific mathematical property of normal distributions in mind and are limited to classify samples by inspecting only that property. The network has more freedom to inspect the sample at hand from different points of view which are determined during training. Also, DBNN is trained on the samples coming from both normal and non-normal distributions. The standard tests calculate their $p$-values based on the normal samples only.  Since DBNN gets to learn the structures of both normal and non-normal samples during the training phase, it can check how well new samples fit within the class of normal as opposed to the class of non-normal distributions. The standard tests, on the other hand, can check only how well a sample fits into a theoretical distribution of a normal distribution's property. This could explain why DBNN was better at detecting non-normality than the tests, but worse at identifying normal distributions. Since the set of rules a network can learn is limited, learning from the non-normal data, which allowed a better performance on the class of non-normal distributions, is likely to have come at the expense of learning from the normal samples in the training dataset. This problem could be overcome by training deeper and more complex networks.

Evaluation on real-world data confirmed DBNN's usefulness in practice. Even when given as few as $5$ numbers from the distribution of earthquake magnitudes, DBNN had the power of almost $100\%$ and was able to detect non-normality in all the samples with $10$ or more elements. The standard statistical tests required a few times larger samples to achieve the same power. When given normal samples of the heights of fewer than $10$ persons, DBNN correctly determined normality in only a little more than $50\%$ of samples. Standard tests beat it in that category. However, on the samples with more than $10$ elements, DBNN correctly detected normality in all but two samples, achieving the power almost equal to $100\%$, which is a result comparable to that of the standard tests at the nominal level of $\alpha=0.01$.

The implications of DBNN's performance are not of interest only to statisticians and the community of machine-learning practitioners and researchers. The size of the sample in each rigorous experiment in science and industry is determined as the minimal size for which the power exceeds the desired level. Given that standard tests lack the power for small samples and that, in some cases, large samples are expensive to obtain or only limited data are available, DBNN can enable experimenters to work with small samples, possibly reducing the cost of data acquisition and allowing for rigorous and valid inference in the cases of small samples, which was previously impossible.

DBNN achieved better results than SBNN, the original approach of \cite{Sigut2006}. This can be explained by different preprocessing steps, as all the other steps in training SBNN were the same as for DBNN. While SBNN completely relies on test statistics and the ways they summarize samples, DBNN mostly uses the empirical quantiles, i.e. the raw data, so it learns classification rules directly from the data instead of the selected summaries that the test statistics represent. 

Although we experimentally determined which quantiles were going to be used, there is still space to study different structures of descriptors. For example, the whole approach may benefit by including some other descriptive statistics, such as sample skewness and kurtosis, or even the very statistics of the standard tests. Although the descriptors that we designed yielded very good results, the optimal structure of the descriptors is yet to be determined. 

The descriptors provide a new way to apply machine-learning algorithms to the samples. Since the samples are sets, which means that they can have any number of elements and that the order of the elements is not relevant, the only way to apply learning algorithms to them was to use set kernels.  However, those kernels can be computationally expensive (e.g. the one from \cite[p. 41]{ShaweTaylor2004} is defined as the number of common subsets between two sets). 

When DBNN was given samples from set $\mathcal{B}$, whose sizes it did not encounter during training, DBNN's performance deteriorated only slightly. Therefore, we argue that the network's performance on the sets of samples with $100$ elements at most should not differ too much from the results that we obtained in our experiments. This should hold because the observed differences in performance were small and the sizes of the samples in $\mathcal{B}$ differ from those of the samples in the training set as much as possible (each size in $\mathcal{B}$, save for the lowest, is the arithmetic mean of two sizes in $\mathcal{A}$).

Even though DBNN was devised as a classifier, it can also quantify uncertainty of its decisions. As noted in Section \ref{sec:neural_networks}, a DBNN's output is actually an estimate of the Bayesian posterior probabilitiy that the input sample is normal. Given that DBNN proved to be fairly calibrated, the probability estimates can be considered reliable. Moreover, the probability of the input sample being normal is easier to understand than the $p$-values with which the standard tests quantify uncertainty and which are frequently misinterpreted. See \cite{Goodman2008} for the details of twelve common fallacies associated with $p$-values.

However, the standard statistical tests do have certain advantages over DBNN. First, their long-term $TPR$ can be set at the level of $1-\alpha$ by choosing $\alpha\in(0, 1)$ as the nominal Type I error rate. We do not have that option with DBNN. The experiments show that DBNN's accuracy is very high for both normal and non-normal distributions and that it increases with the sample size from $n=10$ to $n=100$, however, we cannot control its $TPR$. One way to overcome this shortcoming is to apply appropriate techniques from the Neyman-Pearson classification framework \citep{Dumbgen2008,Tong2013,Tong2018}. Another advantage of the standard tests is that their statistics have mathematical and intuitive interpretations. For example, the statistic of the SW test is a normalized ratio of two estimators of variance: the numerator is the estimator a normal distribution's variance, whereas the denominator is the usual estimator of variance. If the distribution is indeed normal, the ratio should be close to $1$. We do not have that kind of interpretability with DBNN. The network itself with its parameters estimated from the data acts as a statistic, but we cannot assign meaning to its layers and neurons. Its inner dynamics remains largely a black box. Finally, the third advantage of the standard statistical approach is that the power of an established normality test is theoretically guaranteed to reach $1$ for any value of $\alpha$ as the sample size $n$ increases to infinity. Unfortunately, we do not have similar guarantees with DBNN. The network was empirically tested on various and large sets of samples, both real-world and artificial. Its accuracy grew with the sample size. The more similar a distribution is to those we used during training, the higher is the chance that DBNN will correctly classify it as normal or non-normal. Although we used very diverse sets of distributions for training and subsequent evaluation, and made sure that they were large enough so that the estimators of the performance metrics have low variance, we do not have theoretical proofs about DBNN's performance. 

The training of DBNN was be quick and stable, as confirmed by the runtime analysis in Section \ref{subsec:runtime_analysis}.

To get a glimpse at how DBNN fares when given larger samples, we additionally evaluated our network's performance on $7830$ normal and $7830$ non-normal samples with $n=250, 500, 1000$ elements. This set was constructed in the same way as sets $\mathcal{A}$ and $\mathcal{D}$, except that the sample sizes were different. The results are presented in the top part of Table \ref{tab:large_samples}, with the decision threshold equal to the default value of $0.5$. We see that the accuracy on the class of normal distributions drops sharply. DBNN's $TPR$ is only $0.054$ for $n=1000$. However, $AUROC$ is almost $1$ for all three values of $n$, which indicates that DBNN retains its classification abilities for larger samples and that the decision threshold of $0.5$ is not appropriate. When we set it, for each $n$, to the value that corresponds to the point on the respective ROC curve closest to the upper left corner, DBNN exhibited very good performance, with $TPR$ between $0.97$ and $0.99$, and $TNR$ being close to $0.96$. The results are in the bottom part of Table \ref{tab:large_samples}, where those optimal thresholds are also shown. We see that the optimal classification threshold decreases with $n$, which, in turn, implies that the network's estimates of posterior probabilities lose reliability. Those estimates can be made reliable again using some technique of calibration, e.g. Platt Scaling \citep{Platt1999} or isotone regression \citep{Zadrozny2002}, to name just two of them. Another approach would be to train a network on larger samples as well as on the small ones. This additional experiment illustrates that DBNN's $TPR$ may be increased after training by optimizing the decision threshold for each sample size.

\begin{table}
	\centering
	\caption{Performance of DBNN on larger samples (size $n$) with the default ($0.5$) and AUROC-optimized decision thresholds.}
	\label{tab:large_samples}
	\begin{tabular}{lllllllll}
		\toprule
		$n$ & threshold &A &     TPR &     PPV &     TNR &     NPV &      F1 &   AUROC \\
		\midrule
		$250$ & $0.5$ & $0.887$ & $0.779$ & $0.992$ & $0.994$ & $0.818$ & $0.873$ & $0.995$ \\
		$500$  & $0.5$ & $0.766$ & $0.534$ & $0.998$ & $0.999$ & $0.682$ & $0.696$ & $0.994$ \\
		$1000$  & $0.5$ & $0.527$ & $0.054$ & $1.000$ & $1.000$ & $0.514$ & $0.102$ & $0.995$ \\
		\midrule
		$250$     & $0.0635$ & $0.964$ & $0.972$ & $0.957$ & $0.957$ & $0.971$ & $0.964$ & $0.995$ \\
		$500$     & $0.0016$ & $0.968$ & $0.977$ & $0.959$ & $0.959$ & $0.977$ & $0.968$ & $0.994$ \\
		$1000$    & $1.2947\times 10^{-6}$ & $0.976$ & $0.992$ & $0.961$ & $0.960$ & $0.992$ & $0.977$ & $0.995$ \\
		\bottomrule
	\end{tabular}
\end{table}

Proposals for future research are:
\begin{itemize}
	\item finding a way to control DBNN's $TPR$;
	\item training more complex and deeper architectures on possibly larger and more diverse datasets; and
	\item applying the same approach to test for the goodness-of-fit of other distributions beside normal;
\end{itemize}
	
\section{Conclusion}\label{sec:conclusion}

In this paper, we present a neural network that can successfully classify a distribution as normal or non-normal by inspecting a small sample drawn from it. Extensive evaluation on artificial and real-world data show that the network surpasses the standard statistical tests of normality in terms of overall accuracy and power. Although there are ways to improve its performance, the network is a very efficient novel tool for normality testing  as it is. As tests of normality are among the most commonly used statistical tests, the proposed network has a very high potential to be frequently applied in everyday practice of statistics and machine learning in both science and industry. 

\section*{Acknowledgments}
The author would like to thank his advisor, Dr. Milo{\v s} Stankovi{\' c} (Innovation Center, School of Electrical Engineering, University of Belgrade), for useful discussions and advice, and Dr. Wittawat Jitkrittum (Google Research) for advice on the kernel tests of goodness-of-fit. The earthquake data for this study come from the Berkeley Digital Seismic Network (BDSN), doi:10.7932/BDSN, operated by the UC Berkeley Seismological Laboratory, which is archived at the Northern California Earthquake Data Center (NCEDC), doi: 10.7932/NCEDC, and were accessed through NCEDC.
\clearpage
\newpage
\bibliography{literatura}

\end{document}